\RequirePackage[svgnames]{xcolor}

\documentclass[11pt,letterpaper]{mystyle}

\usepackage[all]{hypcap}
\usepackage[svgnames,dvipsnames]{xcolor}
\usepackage[comma,authoryear,compress]{natbib}
\usepackage{algorithm}
\usepackage{algorithmicx}
\usepackage{algpseudocode}
\usepackage{microtype}
\usepackage{graphicx}
\expandafter\def\csname ver@subfig.sty\endcsname{}
\usepackage{booktabs} %
\usepackage{float}
\usepackage{bigstrut}

\usepackage{amsmath}
\usepackage{amssymb}
\usepackage{mathtools}
\usepackage{amsthm}
\usepackage{mathrsfs}
\usepackage{nicefrac}
\usepackage{dsfont}
\usepackage{enumitem}
\usepackage{subcaption}
\usepackage{graphicx,subfig}
\usepackage{cleveref}
\usepackage{bxcoloremoji}

\usepackage{float}

\usepackage[utf8]{inputenc} %
\usepackage[T1]{fontenc}    %
\usepackage{url}            %
\usepackage{booktabs}       %
\usepackage{amsfonts}       %
\usepackage{nicefrac}       %
\usepackage{microtype}      %
\usepackage{graphicx}
\usepackage{subcaption}
\usepackage{amssymb}
\usepackage{fdsymbol}
\usepackage{wrapfig}
\usepackage{lipsum}
\usepackage{enumitem}
\usepackage{stackengine}
\usepackage[font=small,labelfont=bf]{caption}
\usepackage{color}
\usepackage{adjustbox}

\usepackage{rotating}
\usepackage{makecell}

\newcommand{\x}{\mathbf{x}}

\definecolor{blanchedalmond}{rgb}{1.0, 0.92, 0.8}
\definecolor{carmine}{rgb}{0.59, 0.0, 0.09}
\definecolor{lightblue}{rgb}{0.22,0.45,0.70}%

\renewcommand{\mathbf}{\boldsymbol}

\makeatletter
\def\Ddots{\mathinner{\mkern1mu\raise\p@
\vbox{\kern7\p@\hbox{.}}\mkern2mu
\raise4\p@\hbox{.}\mkern2mu\raise7\p@\hbox{.}\mkern1mu}}
\makeatother

\definecolor{amaranth}{rgb}{0.9, 0.17, 0.31}
\definecolor{antiquebrass}{rgb}{0.8, 0.58, 0.46}
\definecolor{antiquefuchsia}{rgb}{0.57, 0.36, 0.51}
\definecolor{chromeyellow}{rgb}{0.31, 0.47, 0.26}

\newcommand{\github}{\raisebox{-1.5pt}{\includegraphics[height=1.05em]{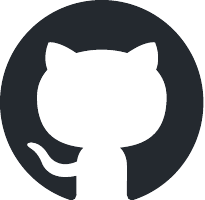}}}

\newtcolorbox{AIbox}[2][]{aibox,title=#2,#1}
\definecolor{lightblue}{rgb}{0.22,0.45,0.70}%
\definecolor{Gray}{gray}{0.95}
\definecolor{Cornsilk}{rgb}{1.0, 0.97, 0.86}

\usepackage{amsmath}

\usepackage[all]{hypcap}

\usepackage{pifont}
\newcommand{\cmark}{\ding{51}} 
\newcommand{\xmark}{\ding{55}} 

\tcbuselibrary{breakable}
\tcbuselibrary{skins}

\newtcolorbox[auto counter, number within=section]{promptbox}[2][]{ 
    colback=white,
    colframe=black,
    arc=5pt,
    outer arc=5pt,
    boxrule=0.8pt,
    fonttitle=\bfseries\sffamily,
    colbacktitle=black,
    coltitle=white,
    enhanced,
    breakable,
    title after break={Prompt \thetcbcounter: #2 (Continued)},
    before skip=10pt plus 2pt,
    after skip=10pt plus 2pt,
    attach boxed title to top left={xshift=5mm, yshift=-3mm},
    boxed title style={
        sharp corners=south,
        arc=2pt,
        boxrule=0.5pt
    },
    title=Prompt \thetcbcounter: #2,
    code={\ifstrempty{#1}{}{\tcbset{label={#1}}}}, 
}

\title{When Structured Sparse Autoencoders Learn Consistent Concepts Across Modalities}

\runningtitle{When Structured Sparse Autoencoders Learn Consistent Concepts Across Modalities}

\author[1]{Weiduo Liao}
\author[2]{Yunqiao Yang}
\author[1]{Ying Wei}

\affil[1]{Zhejiang University}
\affil[2]{Nanyang Technological University}

\correspondingauthor{Ying Wei \href{https://wei-ying.net/}{https://wei-ying.net/}}

\begin{document}

\begin{abstract}
Sparse autoencoders (SAEs) have emerged as a promising technique for mechanistic interpretability by learning a set of sparse latent features in large models, each of which encodes a distinct concept.
However, in vision-language models (VLMs), vanilla SAEs struggle to learn modality-consistent concepts, with concepts often exhibiting fragmented coverage (i.e., disjoint regions) in the visual modality.
To address this challenge, 
we propose 
a \textbf{S}tructured \textbf{S}parse \textbf{A}uto\textbf{E}ncoder (S$^2$AE) that enforces concept consistency from both semantic and spatial perspectives in the visual modality.
Specifically, we group image patches 
based on Transformer attention similarity and spatial proximity, and 
introduce a structured sparsity regularization when training the vanilla SAE.
The regularization consists of 
exclusive sparsity for inter-group concept disentanglement and group sparsity for intra-group concept consistency, which
drives the latent neurons by SAEs to specialize in distinct, semantically grounded concepts. 
Evaluated on the \texttt{Qwen2.5-VL-7B-Instruct} model, the method achieves $6.06\%$ average improvement in semantic alignment (mIoU) and $60.81$ in representational efficiency (lower $l_{0}$ norm) while maintaining near-perfect reconstruction fidelity with an Explained Variance above $99\%$. 
Cross-modal analysis further demonstrates that S$^2$AE enhances neuronal 
monosemanticity
by this visual structural prior, achieving 
a $3.08\%$ average gain in semantic consistency and a $2.37\%$ average gain in monosemanticity scores for both modalities of multimodal features, thereby fostering more coherent and disentangled representations.

\vspace{5mm}


\github{} \textbf{Code Repository}: \href{https://github.com/liaoweiduo/s2ae}{github.com/liaoweiduo/s2ae}.


\coloremojicode{1F917} \textbf{SAE visualization space}: \href{https://huggingface.co/spaces/liaoweiduo/SAE-explorer-dev}{huggingface.co/spaces/liaoweiduo/SAE-explorer}.
\end{abstract}

\maketitle
\vspace{3mm}

\section{Introduction}




Recent large vision-language models (VLMs) have achieved strong performance across diverse vision-language tasks, including image understanding~\citep{hu2022scaling, jung2025visual}, medical image diagnosis~\citep{xiang2025vision, ding2025multimodal}, and increasingly complex multi-modal reasoning~\citep{zhao2025unsupervised, liu2025visual, wang2026vgr}.
However, their increasing deployment has raised concerns regarding the reliability~\citep{guan2024hallusionbench, yang2025mitigating}, generalization~\citep{li2025unveiling, yang2025escaping}, and internal mechanisms~\citep{neo2025towards, jiang2025interpreting} of these models. 
Mechanistic interpretability aims to address these challenges by uncovering the internal representations and computational mechanisms underlying model predictions~\citep{wang2023interpretability, BrickenEtAl2023Monosemanticity, dreyer2025mechanistic}, offering a principled pathway toward understanding, diagnosing, and ultimately improving VLMs in turn~\citep{yao2024knowledge, cywinski2025saeuron, li2026causal}.

\begin{figure}[t]
\centering
\includegraphics[width=\textwidth]{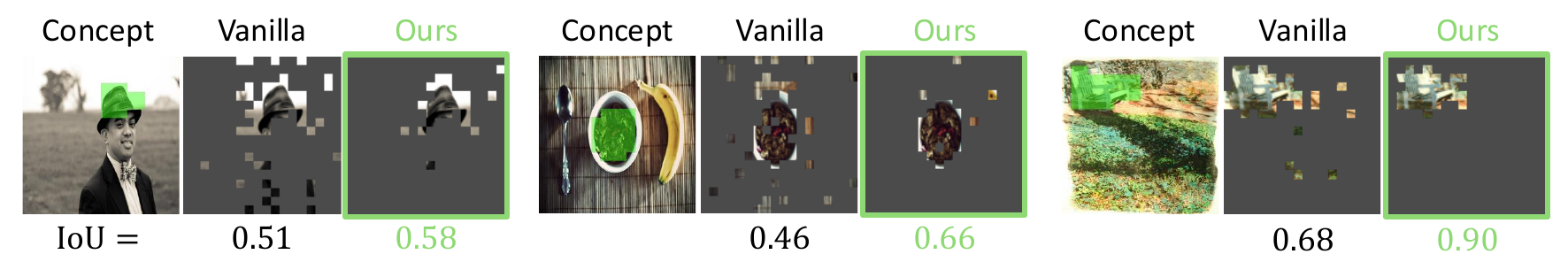}
\caption{
Visualization of SAE activation masks. 
SAE is trained on \texttt{Qwen2.5-VL-7B-Instruct}'s layer 5.  
Three concept cases are shown in \textcolor{ForestGreen}{green} masks on the original images to compare the SAE activation mask performance. 
\textbf{Takeaway:} Our method achieves a significantly closer visual match with the underlying semantic concepts (higher IoU scores with the corresponding concept regions). 
In contrast, the vanilla-learned SAE often exhibits residual polysemanticity.
}
\label{fig:cluster-mask-small}
\end{figure}

Prior neuron-level interpretation of hidden representations~
\citep{zhou2018interpreting, oikarinen2023clipdissect, ahn2024unified} faces a key challenge: individual neurons are often \textit{polysemantic}, activating in response to multiple unrelated concepts~\citep{olah2020zoom, elhage2022toy, gandelsman2025interpreting}. Motivated by the linear representation hypothesis~\citep{park2024linear}, which posits that model representations can be expressed as linear combinations of monosemantic features, sparse autoencoders (SAEs)~\citep{lee2006efficient} project hidden representations into a large concept space that disentangles entangled representations into interpretable conceptual features~\citep{huben2024sparse, lim2025sparse, zhang2025large}. However, vanilla SAEs optimize only for reconstruction fidelity and element-wise sparsity, without explicitly enforcing semantic or spatial coherence among the features. Consequently, the learned conceptual features often exhibit residual polysemanticity, i.e., a single feature responds to multiple unrelated concepts. 
This is reflected in their input activations\footnote{We distinguish between input activations and SAE activations. The former refers to input tokens or image regions that strongly activate a given conceptual feature, while the latter denotes activation values in the SAE layer for a given input (or its subcomponents, e.g., image patches.}, i.e., those input tokens or image regions that activate a concept feature, which are distributed across semantically inconsistent regions~\citep{zaigrajew2025interpreting, han2025causal} rather than aligning with a coherent visual concept. 
For example, a conceptual feature associated with a \texttt{park bench} may also have input activations on \texttt{grass} regions, as illustrated in Figure~\ref{fig:cluster-mask-small}.

To address this issue, we propose a structured sparse autoencoder (S$^2$AE) 
that enforces a visual region to activate semantically coherent 
conceptual features.
A key challenge, however, is that identifying a coherent visual region corresponding to a single concept is itself non-trivial, as concepts are often distributed and not explicitly aligned with pixel-level boundaries. To tackle this, 
we propose to partition each image into semantically coherent visual regions by jointly leveraging Transformer attention similarity and spatial proximity~\citep{haurum2024agglomerative}.
Intuitively, 
Transformer attention captures contextual semantic affinities among patches~\citep{he2022attribute}, while spatial adjacency reflects the structural continuity of visual objects; together 
their combination provides a principled proxy for grouping patches that are likely to share a common underlying concept into a region.
Building on these regions, we introduce \textbf{structured sparsity regularization}, including an \textit{exclusive sparsity} 
regularization that encourages concept features to specialize across different regions and a \textit{group sparsity} regularization that encourages each concept feature to cover tokens throughout a region~\citep{yoon2017combined}. This design promotes inter-region disentanglement and intra-region  
coherence semantically, thereby mitigating residual polysemanticity in the learned concept space.

Importantly, although our structural sparsity regularization is imposed only on visual tokens, its effect is not confined to the visual modality. 
In S$^2$AEs, the training objectives are reconstructing image patches and text tokens through a shared SAE feature space. 
Therefore, when visually fragmented or noisy activations are consolidated into more coherent region-grounded concepts, the corresponding shared SAE features can also acquire a cleaner semantic alignment across modalities. 
In \tableautorefname~\ref{table:modality-summary}, we calculate the monosemanticity (MS) score~\citep{pach2026sparse} for both modalities.
The results show that beyond improving vision-side monosemanticity, S$^2$AE also increases the language-side monosemanticity of multimodal features and improves vision-language consistency.
This suggests that improving visual concept decomposition may indirectly reduce semantic ambiguity in multimodal features and strengthen their alignment with textual activations.

Regarding the monosemanticity evaluation, we also propose 
an automated cross-modal interpretability pipeline to 
evaluate the monosemanticity of
SAE features\footnote{Unless otherwise specified, we use ``SAE features'' and ``conceptual features'' interchangeably.} 
within each modality and their consistency across modalities. 
For each dimension of SAE features, we collect masked images and masked textual contexts that strongly activate the feature, 
and assess whether the semantics inferred from these images and textual contexts are conceptually coherent and consistent across modalities. 
A straightforward approach to autonomously extracting the semantics would directly summarize these multimodal signals using a vision-language model; however, we empirically observe in \sectionautorefname~\ref{sec:exp:pipline_design} that existing VLMs often yield low-quality concept descriptions due to insufficient reasoning reliability.
To bridge this gap, we adopt a hierarchical summarization strategy: we first convert masked images into concise textual descriptions using a vision-language model (i.e., \texttt{Qwen3-VL-8B-Instruct}), and feed these descriptions together with masked textural contexts to 
to a large language model (i.e., \texttt{Qwen3-30B-Instruct}) to (1) generate 
concept explanations for each modality and (2) assess their cross-modal consistency.

Empirically, the proposed structured 
sparsity regularization over SAE features yields concept features that are more visually coherent and semantically interpretable. Evaluated on the Qwen2.5-VL-7B-Instruct model, it improves \emph{monosemanticity} measured by the overlap between 
patches that activate a single conceptual feature 
and a semantically coherent visual region, 
increasing mIoU at layer 15 from 0.516 for (vanilla SAE) to 0.594, while maintaining near-perfect reconstruction fidelity with explained variance above 0.99. Moreover,
the proportion of SAE features with consistent visual and language conceptual explanations rises by 4.3 \% on layer 5, reflecting stronger semantic \emph{consistency} across modalities in the learned concept space.

\section{Related Work}


\paragraph{Vision-Language Models}
Large vision-language models (VLMs) enable cross-modal reasoning by integrating visual representations from vision encoders into large language models, typically through projection layers~\citep{liu2023visual, bai2025qwen3} or cross-attention modules~\citep{alayrac2022flamingo}.
Representative open-source VLM families have evolved along complementary axes: 
InternVL scales from dynamic high-resolution and strong vision encoders toward native multimodal pretraining, RL-based reasoning, and efficient deployment~\citep{chen2024far, wang2025internvl3}; 
Qwen-VL progresses from fine-grained OCR and grounding to dynamic-resolution image/video understanding and MoE-based thinking models~\citep{wang2024qwen2, bai2025qwen3}; 
and GLM-V emphasizes reasoning-centric multimodal training, advancing from compact thinking models to MoE, long-context, and tool-use-oriented variants~\citep{hong2025glm}.
Despite their strong performance across diverse vision-language tasks, the internal representations learned by these models remain highly entangled, making them difficult to interpret mechanistically. 
Recent studies have interpreted VLM representations at multiple levels of granularity, including neurons~\citep{fang2024towards, liu2025modality, xu2025deciphering}, modules~\citep{bi2025unveiling, xia2025one, jiang2025investigating}, and layers~\citep{jiang2025devils, shi2025vision, song2026does}.
Beyond representational analyses, mechanistic interpretability methods have been used to localize functional components through causal tracing~\citep{golovanevsky2024vlms}, activation patching~\citep{liu2026dual, thube2025pathological}, targeted perturbations~\citep{wang2025v}, and attention-head interventions or ablations~\citep{jiang2025investigating}. 
However, these analyses are still typically grounded in architectural units or potentially polysemantic components, rather than fine-grained and interpretable conceptual features.
This limitation motivates feature-level decompositions of multimodal representations into interpretable conceptual features~\citep{lou2025saev, zhang2025large}.

\paragraph{Sparse Autoencoders}
Sparse autoencoders (SAEs) have emerged as a promising tool for mechanistic interpretability by decomposing hidden states into sparse latent features that more closely align with human-interpretable concepts. A range of SAE variants has been proposed to improve feature interpretability, including $\ell_1$-regularized vanilla SAEs~\citep{bricken2023towards}, TopK-based methods~\citep{bussmann2024batchtopk, gao2025scaling}, Gated SAEs~\citep{rajamanoharan2024improving}, and JumpReLU SAEs~\citep{rajamanoharan2024jumping}. These methods have supported a broad range of interpretability analyses and downstream applications across large language models~\citep{gao2024scaling, makelov2025towards}, vision-language models~\citep{lim2025sparse, lou2025saev, zhang2025large, shen2025vlsae, kaushik2026learning}, and cross-model settings~\citep{thasarathan2025universal, sarvi2026sparc}. However, the training objectives of  existing SAE formulations typically 
include reconstruction fidelity and sparsity only, without explicitly modeling 
coherence among concept features. As a result, the concept features still present residual polysemantic~\citep{makelov2025towards, leask2025sparse, chanin2025a, paulo2026sparse}. In the VLM setting, this limitation is particularly evident when input image patches that activate a conceptual feature 
fail to align consistently with a semantically coherent visual region, underscoring the need to leverage 
visual region structures in turn to improve SAE learning.
In contrast to prior multi-modal SAE studies that mainly analyze visual and textual activations after training, our work studies whether structural priors imposed on the visual side can actively shape the shared multi-modal feature space, leading not only to visually coherent features but also to improved cross-modal semantic consistency.

\section{Preliminaries}


\subsection{Sparse Autoencoder}
Consider 
concept extraction of an input \(\mathbf{x}\in\mathbb{R}^{D}\) as a by-product of an unsupervised reconstruction task with sparse auto-encoders~\citep{BrickenEtAl2023Monosemanticity}. 
Examples of \(\mathbf{x}\) include the residual stream representations of any transformer layer from image patches~\citep{lim2025sparse}, text tokens~\citep{lieberum-etal-2024-gemma}, or both~\citep{zhang2025large}.
In our implementation, we explore multi-modal SAEs, where image patches and text tokens are treated as equivalent discrete units during the reconstruction process.
The basic formulation for the encoder and the decoder is as follows:
\begin{equation}
\label{equ:basic-sae}
    \begin{aligned}
        \mathbf{z} &= \sigma((\mathbf{x}-\mathbf{b}_{\mathrm{dec}})\mathbf{W}_{\mathrm{enc}}+\mathbf{b}_{\mathrm{enc}}), \\
        \hat{\mathbf{x}} &= \mathbf{z}\mathbf{W}_{\mathrm{dec}}+\mathbf{b}_{\mathrm{dec}},
    \end{aligned}
\end{equation}
where \(\mathbf{z}\in\mathbb{R}^{N}\) denotes the SAE 
feature activation vector, also known as SAE latent~\citep{zhang2025large} or code~\citep{fel2025archetypal}, \(\mathbf{W}_{\mathrm{enc}}\in\mathbb{R}^{D\times N}\) is the encoder mapping with bias \(\mathbf{b}_{\mathrm{enc}}\in\mathbb{R}^{N}\), and \(\mathbf{W}_{\mathrm{dec}}\in\mathbb{R}^{N\times D}\) is the decoder mapping with bias \(\mathbf{b}_{\mathrm{dec}}\in\mathbb{R}^{D}\). \(N,D\) denote the number of SAE features and the hidden dimensionality of the 
input $\mathbf{x}$, respectively. 
The \(\sigma(\cdot)\) is an activation function (e.g., ReLU) and depends on the specific SAE methods. 
The vanilla training objective in \citet{BrickenEtAl2023Monosemanticity} is to minimize the reconstruction mean square error with an additional sparsity 
loss as follows: \(\mathcal{L}_{\mathrm{vanilla}}=\|\mathbf{x}-\hat{\mathbf{x}}\|_2+\lambda\|\mathbf{z}\|_1\), where \(\|\cdot\|_1\) denotes a \(l_1\) penalty and \(\lambda\) is a coefficient. 
The overwhelmingly large latent dimensionality \(N \gg D\) and 
the sparsity 
regularization
 empowers decomposition of potential monosemantic concepts in \(\mathbf{x}\). 

Top-K SAE~\citep{gao2024scaling} implements the sparsity regularization 
by a hard Top-K mask (i.e., \(\sigma(\cdot)=\text{TopK}(\cdot)\)) on SAE features, avoiding the use of the \(l_1\) penalty
which introduces a shrinkage bias on the SAE activation magnitudes.
To improve SAE feature utilization, Top-K SAE
initializes \(\mathbf{W}_{\mathrm{dec}}\) to the transpose of \(\mathbf{W}_{\mathrm{enc}}\), and uses an auxiliary loss \(\mathcal{L}_{\mathrm{aux}}\) with coefficient \(\lambda_{\mathrm{aux}}\) to reconstruct \(\mathbf{x}\) with top-\(k_{aux}\) SAE features that never being activated so far, as introduced in~\citep{gao2024scaling}. 
Thus, the loss for Top-K SAE is as follows:
\begin{equation}
    \mathcal{L}_{\mathrm{sae}}=\|\mathbf{x}-\hat{\mathbf{x}}\|_2 + \lambda_{\mathrm{aux}}\mathcal{L}_{\mathrm{aux}}.
\end{equation}

\subsection{Group Sparsity and Exclusive Sparsity}
Group sparsity and exclusive sparsity~\citep{yoon2017combined} have been 
two complementary regularization terms 
on a given matrix $\mathbf{Z} \in \mathbb{R}^{M \times N}$, where $M$ and $N$ denote the dimensions of rows and columns, respectively. 
Following the principles of mixed-norm regularization, 
they have been defined in~\citet{yoon2017combined} as follows:

\paragraph{Group Sparsity (GS)}
Group sparsity, typically formulated via the $\ell_{2,1}$ norm, treats each column of a matrix as a group and 
encourages sparsity at the group level, effectively driving 
entire columns to zero. 
For the matrix $\mathbf{Z}$, the group sparsity regularization is defined as 
\begin{equation}
\label{equ:gs-general}
    \mathcal{L}_{\mathrm{gs}}(\mathbf{Z}) = \|\mathbf{Z}\|_{2,1} = \sum_{j=1}^{N} \left( \sum_{i=1}^{M} \mathbf{Z}_{ij}^2 \right)^{1/2},
\end{equation}
where the inner $\ell_2$ norm measures the magnitude of each column, and the outer $\ell_1$ norm promotes sparsity across columns. By penalizing the sum of column magnitudes, GS facilitates column-wise selection, ensuring that only a small subset of columns (i.e., groups) remains active.

\paragraph{Exclusive Sparsity (ES)}
In contrast to group sparsity, exclusive sparsity (typically formulated via the $\ell_{1,2}$ norm) 
encourages 
competition within predefined groups to enforce mutually exclusive 
representations. 
By treating each column of $\mathbf{Z}$  again as a group, the regularization is defined as
\begin{equation}
\label{equ:es-general}
    \mathcal{L}_{\mathrm{es}}(\mathbf{Z}) = \sum_{j=1}^{N} \left( \sum_{i=1}^{M} |\mathbf{Z}_{ij}| \right)^2,
\end{equation}
where the inner $\ell_1$ norm promotes sparsity with each column, and the outer $\ell_2$ norm balances the overall magnitude across columns. 
This forces each group to develop a unique profile, preventing different columns from sharing the same active row indices.

While GS uses an inner $\ell_2$ norm and outer $\ell_1$ norm to achieve ``all-or-nothing'' selection for entire columns, ES utilizes an inner $\ell_1$ norm and outer $\ell_2$ norm to promote ``one-of-many'' behavior within columns. In \sectionautorefname~\ref{sec:esgs}, we will further detail how these 
regularizers are specifically adapted to the activation matrix of our SAE features to achieve semantic monosemanticity and consistency. 

\section{Improving Monosemanticity of Conceptual Features}



While 
SAEs applied on VLMs
provide a framework for unifying vision and language concept 
extraction, 
we empirically observe that visual tokens are discretized from continuous image patches. 
We provide \figureautorefname~\ref{fig:cluster-mask-small}, which is a 
visualization of SAE activation masks 
on three example images, using an SAE trained on layer 5 of \texttt{Qwen2.5-VL-7B-Instruct}. 
For each visual region describing a concept in an image, we identify the specific neuron whose activation mask (comprising all patches that activate it) yields the maximum Intersection over Union (IoU) with this region. 
For the details of this experiment and more examples, please refer to \sectionautorefname~\ref{sec:improved_vision_repr} and \figureautorefname~\ref{fig:cluster-mask}. 
In \tableautorefname~\ref{table:modality-summary}, we report the monosemanticity score~\citep{pach2026sparse} w.r.t. vision and language modalities. 
Vision monosemanticity scores (e.g., $0.492$ for Layer 15) are generally lower than language monosemanticity scores (e.g., $0.869$ for Layer 15), suggesting that learning monosemantic concepts in the vision modality is inherently more difficult for SAEs.

The visualization and quantitative results highlight a fundamental limitation in current SAEs in the vision modality. 
Vanilla SAEs treat image patches identically to text tokens, lacking any structural guarantee for visual concepts. 
As observed in the ``Vanilla'' column, this token-centric approach often leads to significant residual polysemanticity. 
For instance, a SAE feature intended to represent a specific object may erroneously be activated on unrelated patches across the image simply because they share low-level feature similarities, failing to respect the spatial and semantic boundaries inherent to visual data.
This motivates our proposed framework for improving the monosemanticity of conceptual features, moving beyond token-wise sparsity objectives that treat visual patches independently and overlook the spatial correlations inherent in visual representations, toward a formulation that better captures the integrated nature of visual intelligence.
By introducing structural regularizers, we ensure that the extracted SAE features are not only sparsely activated but also spatially and conceptually coherent. 

Moreover, since the SAE is trained over both image patches and text tokens with a shared dictionary, visual and textual activations are not represented by disjoint SAE feature sets. 
A SAE feature that appears multimodal can be activated by both visual regions and textual contexts. 
Thus, visual-side polysemanticity can contaminate the shared semantic identity of such a feature. 
That is, if the visual activations mix object, background, and texture cues, the corresponding textual activations may also become harder to summarize as a single coherent concept. 
Our structured visual regularization targets this source of ambiguity by forcing visual activations to align with coherent regions, which in turn can sharpen the shared multimodal semantic features.

\subsection{Visual Regions from Patch Grouping}
\label{sec:concept_extraction_clustering}
Enabling structured sparsity over SAE features requires a representation of visual regions that serves as a proxy for visually grounded concepts. Such concepts include (i) object-centric concepts (e.g., \texttt{people}, \texttt{tree}) that correspond to concrete entities, and (ii) relation-based or abstract concepts (e.g., \texttt{throwing}, \texttt{happy}) that are grounded in specific visual cues such as body parts or facial expressions.
To extract such regions, we leverage the self-attention mechanism, which characterizes the asymmetric semantic dependencies between patches 
to group semantically related patches. 
Particularly, we adopt 
agglomerative clustering~\citep{haurum2024agglomerative} rather than  K-means~\citep{vyas2020fast}, as its bottom-up merging process better captures the hierarchical structure of concepts, progressively aggregating local textures into higher-level semantic objects. 
In the following, we construct an adjacency matrix \(\mathbf{d}\) that encodes pairwise affinities between patches, combining Transformer attention similarity and spatial proximity, and defines the connectivity used for agglomerative clustering.

\paragraph{Grouping using attention}

Patches belonging to the same concept 
tend to have similar representations and exhibit similar attention patterns in the forwarding process of transformer layers. 
Thus, we leverage this prior assumption to define pairwise affinities between image patches for region grouping.
For a given image, 
let \(\mathbf{Z}^{l}\in\mathbb{R}^{M_I\times N}\) denote the SAE activations extracted at 
transformer layer \(l\), where \(M_I\) is the number of image patches. 
We obtain the corresponding attention matrix \(\mathbf{A}^l\in\mathbb{R}^{M_I\times M_I}\), where each entry reflects the attention weight between pairs of patches. 
To obtain a symmetric adjacency matrix \(\mathbf{d}\) for clustering, we convert such single-directional similarities to bi-directional distances \(\mathbf{d}^A\in\mathbb{R}^{M_I\times M_I}\) as follows: 
\begin{equation}
    \begin{aligned}
        \mathbf{d}^A = \text{Norm}(-(\mathbf{\bar{A}} +\mathbf{\bar{A}}^\top)), 
    \end{aligned}
\end{equation}
where \(\mathbf{\bar{A}}=\frac{1}{L}\sum_{l\in[1,\dots,L]}\mathbf{A}^l\) is the average attention over \(L\) layers, and \(\text{Norm}(\cdot)\) denotes row-wise min-max 
normalization 
to rescale distances in \(\mathbf{d}\) 
to \([0,1]\). 
Note that we use the average attention over all layers (i.e., \(\bar{\mathbf{A}}\)) instead of that on the current layer (i.e., \(\mathbf{A}^{l}\)) for robust clustering. 
For further consideration of this design choice, please refer to \sectionautorefname~\ref{sec:exp:clustering_design}. 

\paragraph{Incorporating spatial matching}
\begin{figure}[ht]
\centering
\begin{subfigure}[b]{0.32\textwidth}
    \centering
    \includegraphics[width=\textwidth]{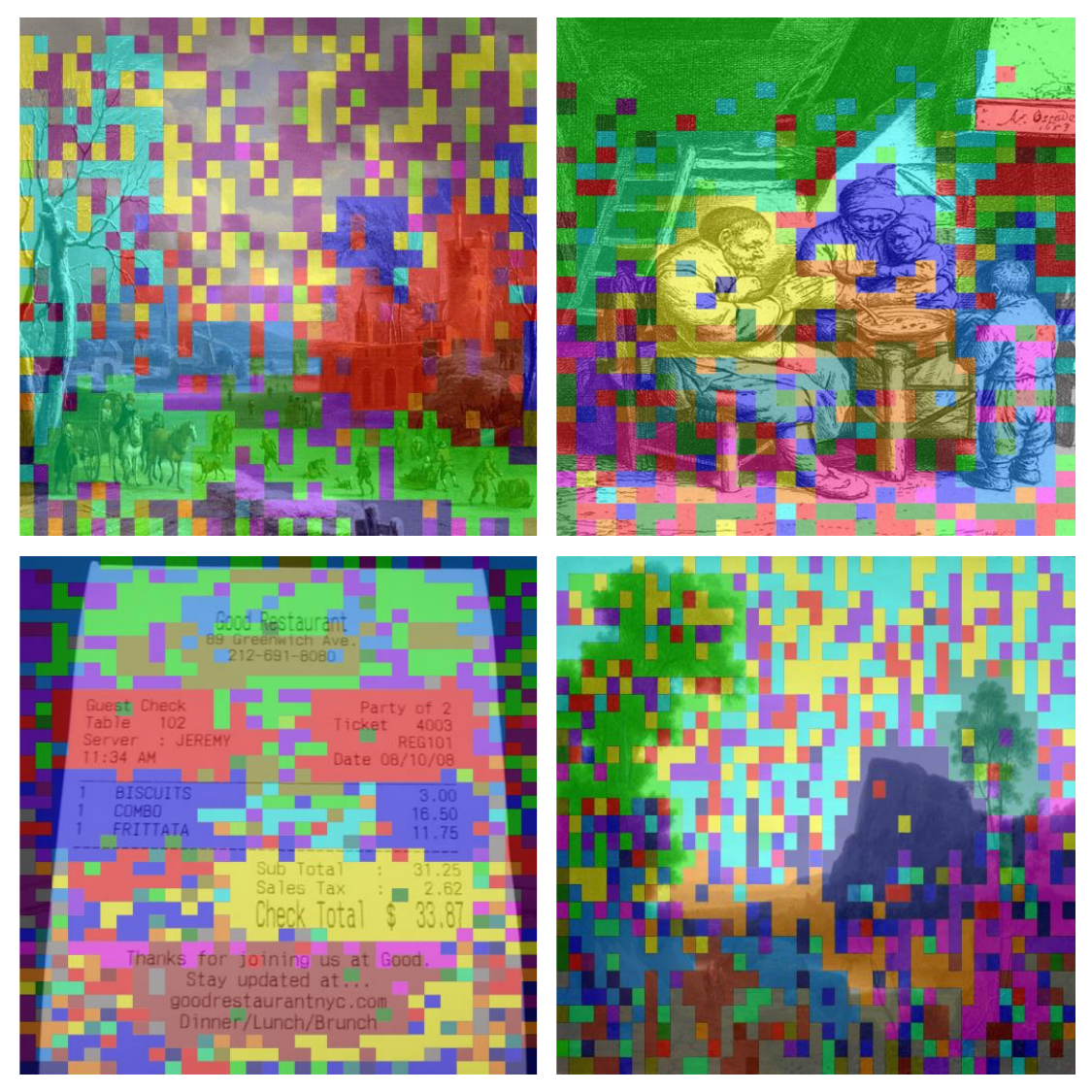}
    \caption{Only attention}
    \label{fig:clustering:attn}
\end{subfigure}
\hfill
\begin{subfigure}[b]{0.32\textwidth}
    \centering
    \includegraphics[width=\textwidth]{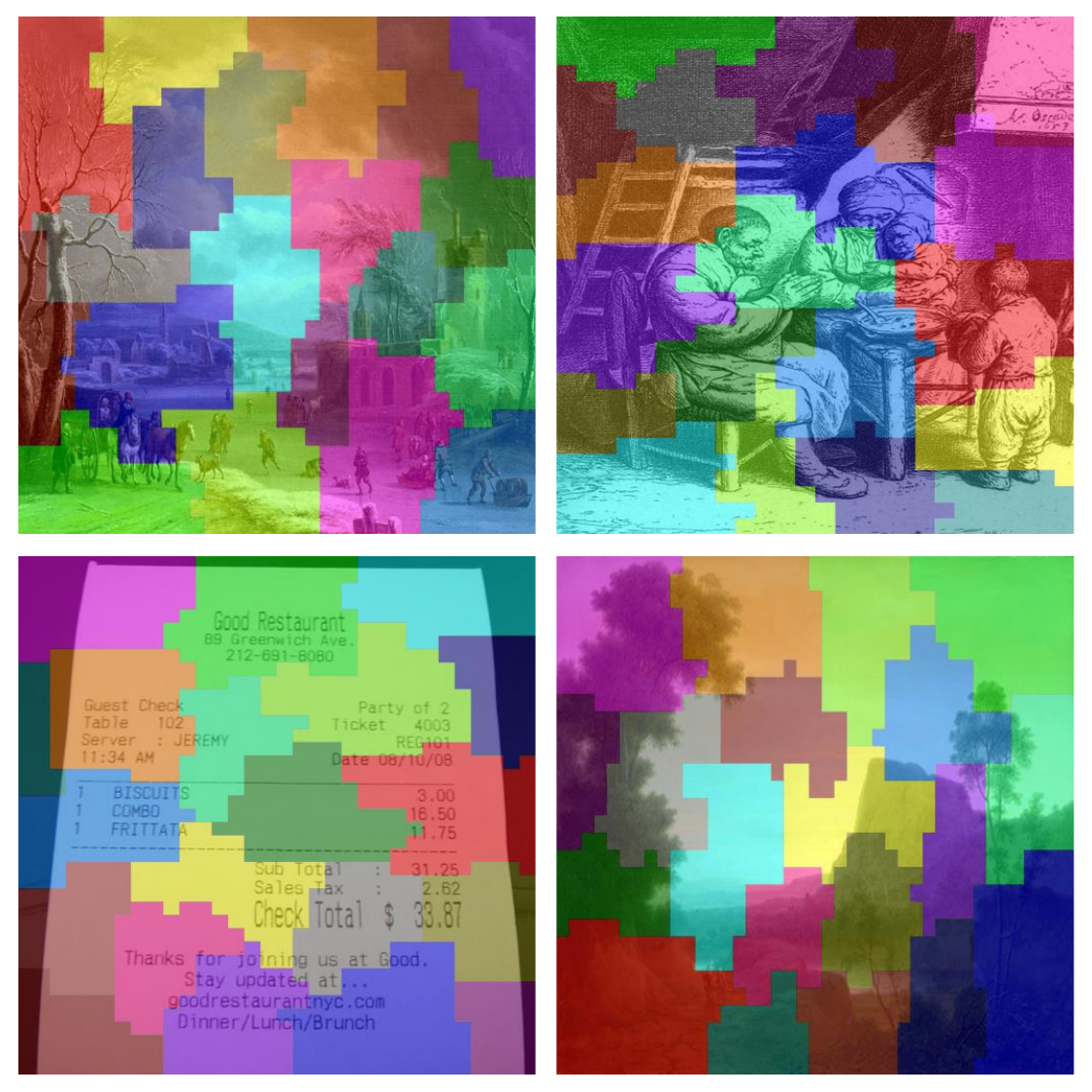}
    \caption{Only spatiality}
    \label{fig:clustering:spatial}
\end{subfigure}
\hfill
\begin{subfigure}[b]{0.32\textwidth}
    \centering
    \includegraphics[width=\textwidth]{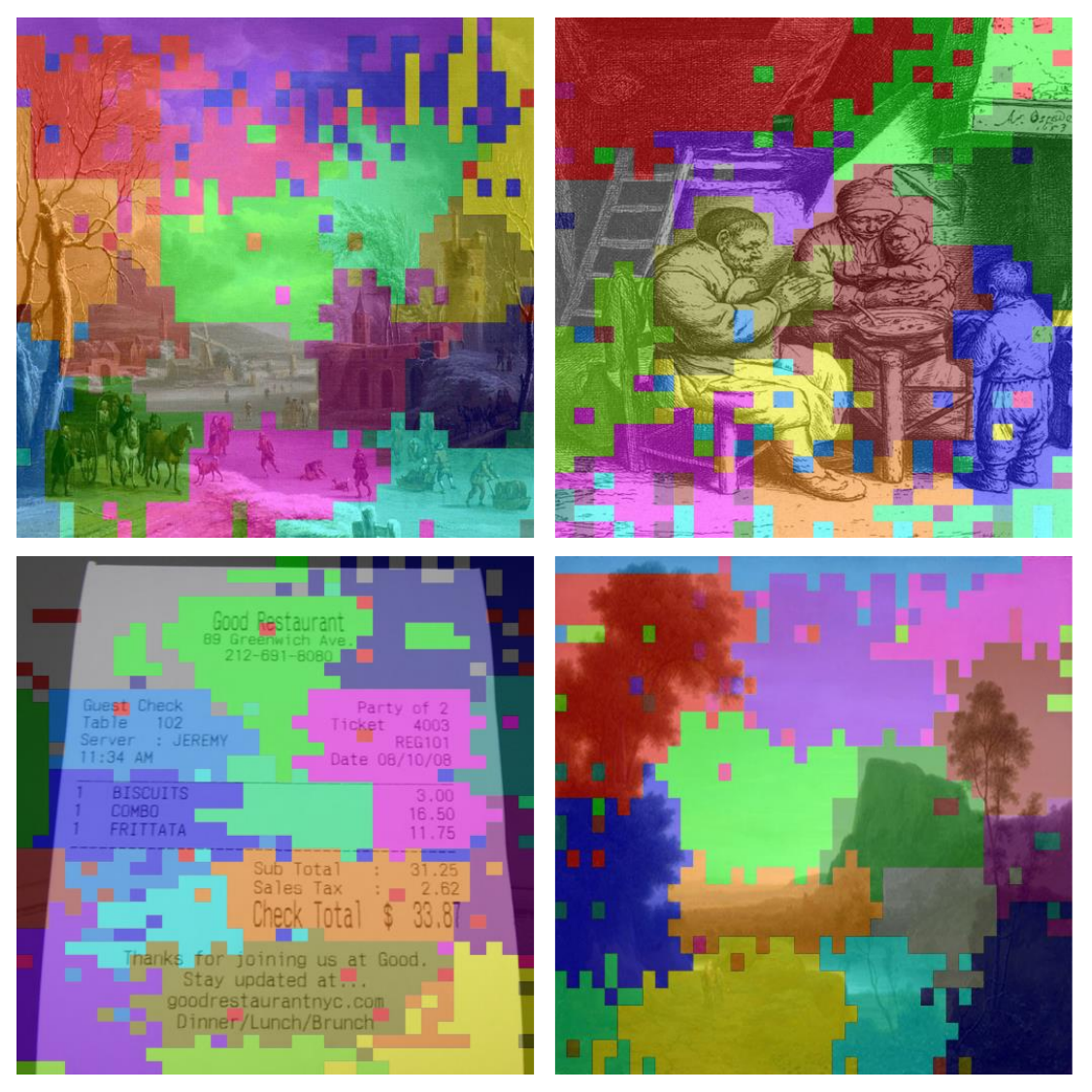}
    \caption{Attention + spatiality}
    \label{fig:clustering:attn+spatial}
\end{subfigure}
\caption{
Visualization of clustering according to different affinity characterization: a) only the attention similarity, b) only the spatial distance, and c) both attention and spatial distance are used to construct the adjacency matrix. 
}
\label{fig:clustering}
\end{figure}

However, 
clustering based solely on attention 
introduces noisy outliers, 
as 
illustrated in
\figureautorefname~\ref{fig:clustering:attn}.
Although 
the grouped visual regions show dominant concepts,
such as \texttt{peoples}, \texttt{trees}, and \texttt{paragraphs}, there exist many patch outliers within a group that are far away from the clustering centers and semantically inconsistent with the dominant concepts. 
This mismatch between attention-based affinity and spatial continuity in natural images, i.e., visual concepts are typically spatially contiguous with patches belonging to the same object forming connected regions, motivates the need for an explicit spatial prior.
Hence, we incorporate the spatial patch distance matrix 
to modulate \(\mathbf{d}^A\).
Specifically, 
we define a spatial distance matrix $\mathbf{d}^S_{i,j}$ based on patch coordinates.
Let the image be divided into \(h\times w\) 
patches with \(hw=M_I\), and let \(\mathbf{p}_{i}\in[1,\dots,h]\times[1,\dots,w]\) denote the coordinates of patch \(i\).
The patch spatial distance matrix \(\mathbf{d}^S\in\mathbb{R}^{M_I\times M_I}\) is defined using the patch-level Manhattan distance~\citep{suwanda2020analysis}:
\begin{equation}
    \begin{aligned}
        \mathbf{d}^S_{i,j} =||\mathbf{p}_i-\mathbf{p}_j||_1, \quad
        \mathbf{d}^S \leftarrow \text{Norm}(\mathbf{d}^S),
    \end{aligned}
\end{equation}
where \(\mathbf{d}^S_{i,j}\) is the element in the \(i\)-th row and the \(j\)-th column of \(\mathbf{d}^S\). 

The final adjacency matrix \(\mathbf{d}\) for clustering is specified as: 
\begin{equation}
\label{equ:adjacency_matrix}
    \begin{aligned}
         \mathbf{d}_i = \mathbf{d}^A_i\odot{(\mathbf{d}^S_i)}^\alpha,
    \end{aligned}
\end{equation}
where \(\alpha\) is a coefficient to balance the impact of spatial distance, which is empirically studied in \sectionautorefname~\ref{sec:exp:clustering_design}. We then perform agglomerative clustering using \(\mathbf{d}\) as the distance matrix to partition the image into $G$ visual regions, each of which characterizes a concept.

\subsection{Exclusive Sparsity and Group Sparsity}
\label{sec:esgs}
To 
encourage each SAE feature 
to specialize in 
a single visual concept, we adapt 
the exclusive sparsity regularization in \equationautorefname~\ref{equ:es-general} 
 to operate over visual regions extracted in \sectionautorefname~\ref{sec:concept_extraction_clustering}.
In the original formulation, exclusive sparsity is applied to a matrix where each column is treated as a group, enforcing competition within each column.
In our setting, we instantiate a group as a visual region that encompasses a group of patches.
To this end, we first construct a group-level SAE activation matrix. 
Let $\mathbf{Z}^{g} \in \mathbb{R}^{M_g \times N}$ be the SAE activations restricted to 
the $g$-th group, where $M_g$ is the number of patches within that group and $N$ is the number of SAE features. 
We 
develop a 
group-level
activation profile vector 
by applying an $\ell_2$-norm across the patch dimension, yielding $\mathbf{s}^g \in \mathbb{R}^N$ such that its $j$-th SAE activation is as follows,
\begin{equation}
s^g_j = \|\mathbf{Z}_{:,j}^{g}\|_2.    
\end{equation}
The $\ell_2$-norm ensures that the vector $\mathbf{s}^g$ reflects the aggregate magnitude of 
SAE
activations within the $g$-th group, providing a differentiable and representative basis for the subsequent exclusive sparsity penalty.
Stacking
these group-level profile vectors 
gives a global activation matrix $\mathbf{S} = [\mathbf{s}^1, \dots, \mathbf{s}^G]^\top \in \mathbb{R}^{G \times N}$, where each row corresponds to a visual region and each column corresponds to a SAE feature.
We then apply the exclusive sparsity formulation in~\equationautorefname~\ref{equ:es-general} to $\mathbf{S}$, yielding
\begin{equation}
    \mathcal{L}_{\mathrm{es}} = \frac{1}{N}\sum_{j=1}^N (\sum_{g=1}^G |s^g_j|)^2.
\end{equation} 
Under this formulation, the $\ell_1$ aggregation over visual groups encourages competition among groups for each SAE feature, while the outer $\ell_2$ norm penalizes dispersed activations across multiple groups. Each SAE feature is thus encouraged to be activated predominantly within a single visual group, reducing cross-group co-activation and improving monosemanticity.

To enforce consistent SAE activation within each visual region as a group, 
we further impose a group 
sparsity regularization. 
The goal is to encourage patches within the same group to activate a shared, sparse subset of SAE features, thereby improving intra-group semantic coherence.
We adapt the group sparsity formulation in~\equationautorefname~(\ref{equ:gs-general}) to operate on 
the previously defined group-level activation profile vectors $\mathbf{s}^g$; 
we apply an $\ell_1$ norm across the SAE feature dimension (i.e., $N$) for each group:
\begin{equation}
    \mathcal{L}_{\mathrm{gs}} = \frac{1}{G} \sum_{g=1}^G \|\mathbf{s}_g\|_1.
\end{equation}
By minimizing this objective, we achieve higher intra-group monosemanticity, 
where each visual region is represented by a more semantically consistent and sparse set of SAE features.

To optimize the SAE under these structural regularizations, we augment the original SAE objective with the two proposed regularizers. 
The total loss function is defined as follows: 
\begin{equation}
    \mathcal{L}_{\mathrm{total}} = \mathcal{L}_{\mathrm{sae}} + \lambda_{\mathrm{es}} \mathcal{L}_{\mathrm{es}} + \lambda_{\mathrm{gs}} \mathcal{L}_{\mathrm{gs}},
\end{equation}
where $\lambda_{\mathrm{es}}$ and $\lambda_{\mathrm{gs}}$ are hyperparameters that govern the trade-off between 
reconstruction fidelity and structural regularization. 
Specifically, the exclusive sparsity term $\mathcal{L}_{\mathrm{es}}$ encourages inter-group competition for SAE feature specialization, while the group sparsity term $\mathcal{L}_{\mathrm{gs}}$ promotes intra-group coherence by aligning 
SAE activations across patches within the same group.
Together, these complementary regularization 
ensure 
the resulting SAE features to be not only sparse but also semantically 
aligned with the 
underlying visual region structure. 

\paragraph{Binarized SAE activations}
To further decouple feature selection from activation magnitude and mitigate the ``shrinkage bias'' commonly observed in $\ell_1$-regularized models, we introduce a binarization layer with a Straight-Through Estimator~\citep{bengio2013ste} to the latent representations. 
Specifically, the continuous activations $\mathbf{Z}^{g}$ are transformed into binary gates $\hat{\mathbf{Z}}^{g}$ using the formulation $\hat{\mathbf{Z}} = \text{bin}(\mathbf{Z}) + \mathbf{Z} - \mathbf{Z}.\text{detach()}$. 
Consequently, the subsequent structural penalties, including $\mathcal{L}_{\mathrm{es}}$ and $\mathcal{L}_{\mathrm{gs}}$, are computed based on these binary gates rather than 
raw SAE activation magnitudes. 
This modification ensures that the sparsity regularization specifically target 
activation frequency and group-wise SAE feature co-occurrence, 
rather than directly penalizing SAE feature intensities.

\section{Interpreting SAE Features via Hierarchical Semantic Synthesis}
This section details our hierarchical framework for interpreting the SAE features learned by S$^2$AEs.
The goal is to move beyond mere activation visualization toward a rigorous, automated understanding of how individual SAE features encode concepts across both visual and textual domains.

\subsection{Motivation}
Interpreting SAE features in VLMs requires assigning a concise semantic explanation to the inputs (a.k.a. \texttt{references}) that strongly activate each SAE feature. 
Unlike language-only SAEs, where activating tokens are already expressed in natural language, visual SAE activations are masked image regions~\citep{bohacek2026uncovering}. 
Pioneered by~\citet{bills2023language,zhang2025large,gandelsman2024interpreting}, these regions can be partial, small, spatially fragmented, or mixed with nearby background cues. 
Therefore, a reliable interpretation pipeline must first identify what each activated region depicts before abstracting a shared concept across multiple references.

A straightforward solution is direct summarization: given a set of top-activating image/text references, directly ask a VLM to summarize their common visual/language concept. 
However, this formulation couples several challenges into a single step, like recognizing masked visual evidence, filtering noisy or incomplete activations, comparing multiple references, and abstracting their shared semantics. 
Empirically, we find that this direct pipeline often fails to produce valid summaries, especially for the vision modality. 
As shown in \sectionautorefname~\ref{sec:exp:pipline_design}, its identification rate (e.g., 66.4\% at Layer 5) is substantially lower than our pipeline (e.g., 98.8\% at Layer 5) across all evaluated layers. 
This suggests that current VLMs are not sufficiently reliable when asked to perform direct cross-reference concept abstraction from masked activations.

\begin{figure}[ht]
\centering
\includegraphics[width=\textwidth]{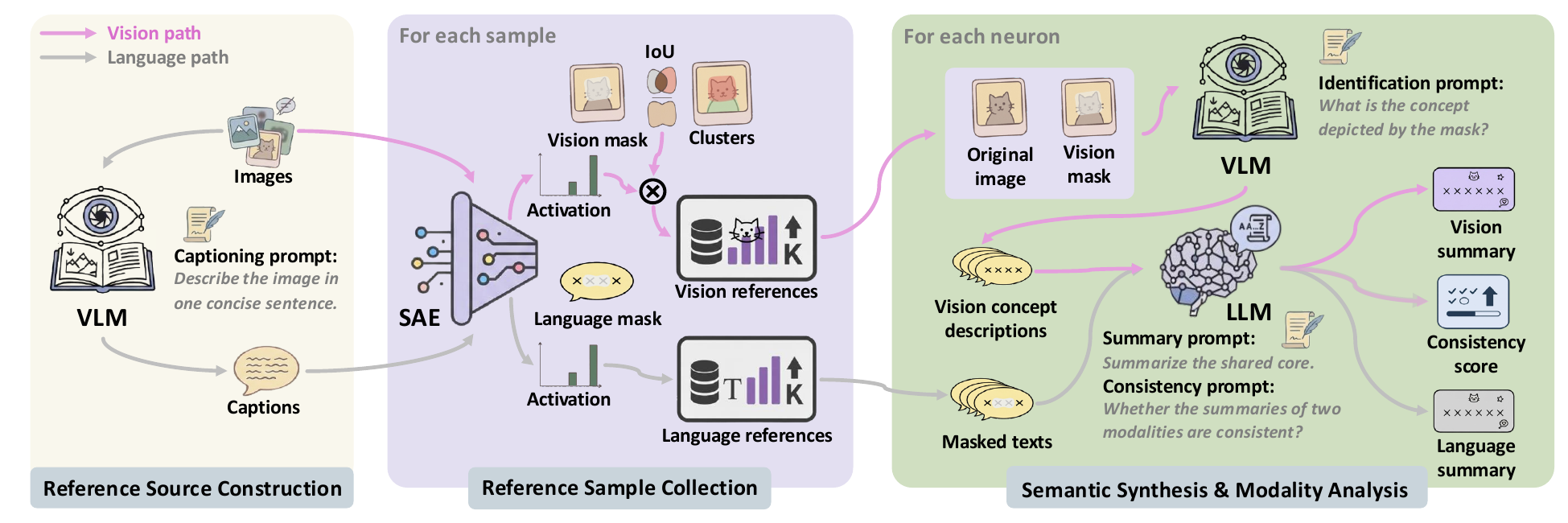}
\caption{
Overview of the SAE feature summarization pipeline.
\textbf{Reference Source Construction:} The pipeline begins by using a VLM to generate concise captions for raw images, creating a paired multi-modal dataset as the foundation for analysis.
\textbf{Reference Sample Collection:} Vision reference samples are prioritized by a composite score, which is the product of the activation and the maximum IoU between the SAE feature's activation mask and semantic clusters, ensuring that selected samples are both highly active and conceptually coherent, while language reference samples only depend on activation values. 
\textbf{Semantic Synthesis:} Specifically for visual samples, a VLM serves as a high-level annotator to translate the isolated visual concepts within the masks into natural language concept descriptions. 
\textbf{Modality Analysis:} After that, an LLM independently processes the textual descriptions from the vision path and the raw masked tokens from the language path to summarize the core semantic themes of each modality.
Finally, the LLM compares the two modality-specific summaries to calculate a consistency score, determining whether the SAE feature represents a unified concept across both vision and language.
Section~\ref{sec:exp:pipline_design} further compares this hierarchical design with a direct VLM summary baseline.
}
\label{fig:ref-summary}
\end{figure}

We therefore adopt a hierarchical interpretation pipeline in \figureautorefname~\ref{fig:ref-summary}. 
For each SAE feature, we first perform reference-level concept identification. 
Masked image references are converted into concise visual descriptions by a VLM, while masked text references are converted into concise textual descriptions by an LLM. 
We then perform SAE feature-level semantic synthesis, where an LLM summarizes the visual descriptions and textual descriptions into modality-specific concepts and compares them for cross-modal consistency. 
This decomposition reduces the reasoning burden of each model call. 
The VLM focuses on localized visual recognition, while the LLM handles abstraction and consistency judgment in a unified textual space. 
As a result, the pipeline provides more stable feature summaries and a more reliable basis for measuring modality 
, cross-modal consistency, and monosemanticity.

\subsection{Reference Source Construction}
\label{sec:reference-source-construction}
To perform feature-level interpretation of our S$^2$AE, we require a comprehensive set of reference samples where each image is paired with a descriptive text counterpart. 
Our data pipeline consists of two primary phases: 1) sourcing high-quality visual data and 2) generating synthetic textual descriptions to achieve cross-modal alignment.

\paragraph{Visual data sourcing}
For the image modality, we utilize the evaluation dataset provided by~\citet{zhang2025large}.
This dataset, hosted on \texttt{Hugging Face}
\footnote{Hugging Face website: \href{https://huggingface.co/datasets/lmms-lab/sae-sample-cache-dataset}{https://huggingface.co/datasets/lmms-lab/sae-sample-cache-dataset}.}, serves as an extensive repository of visual stimuli curated for interpretability research. 
While this source provides a diverse range of images, it does not include a corresponding high-granularity text corpus for every sample.

\paragraph{Synthetic captioning via VLM}
To bridge the gap between modalities and create a balanced multi-modal reference set, we implement a customized textual augmentation pipeline. 
Given the absence of native text data for the aforementioned image source, we utilize \texttt{Qwen3-VL-8B-Instruct}~\citep{qwen3technicalreport} to generate high-fidelity synthetic captions, using the prompt: \textit{Describe the image in one concise sentence}.
This augmented dataset allows us to collect reference samples for each learned SAE feature across both modalities. By identifying the top-activating samples for a specific SAE feature, we can analyze the visual and textual activations. 

\subsection{Reference Sample Collection and Masking}
To interpret the learned SAE features, we follow a rigorous sample collection process, inspired by~\citet{zhang2025large}. 
For each SAE feature, we identify some top activating samples from a large-scale multimodal corpus. 
However, raw activations often contain noise from irrelevant spatial or linguistic contexts.
Thus, we employ a masking strategy to achieve precise attribution. 
For each reference sample, including the image and its captioning text, we apply activation masks, retaining only the patches/tokens that have positive activation values for the SAE feature, for both vision and language modality.
This masking approach ensures that our interpretation is grounded in the specific sub-components of the input that the SAE feature has specialized in, rather than the global context of the sample.

\paragraph{Activation-semantic alignment for higher-quality reference samples} 

\begin{figure}[ht]
\centering
\includegraphics[width=\textwidth]{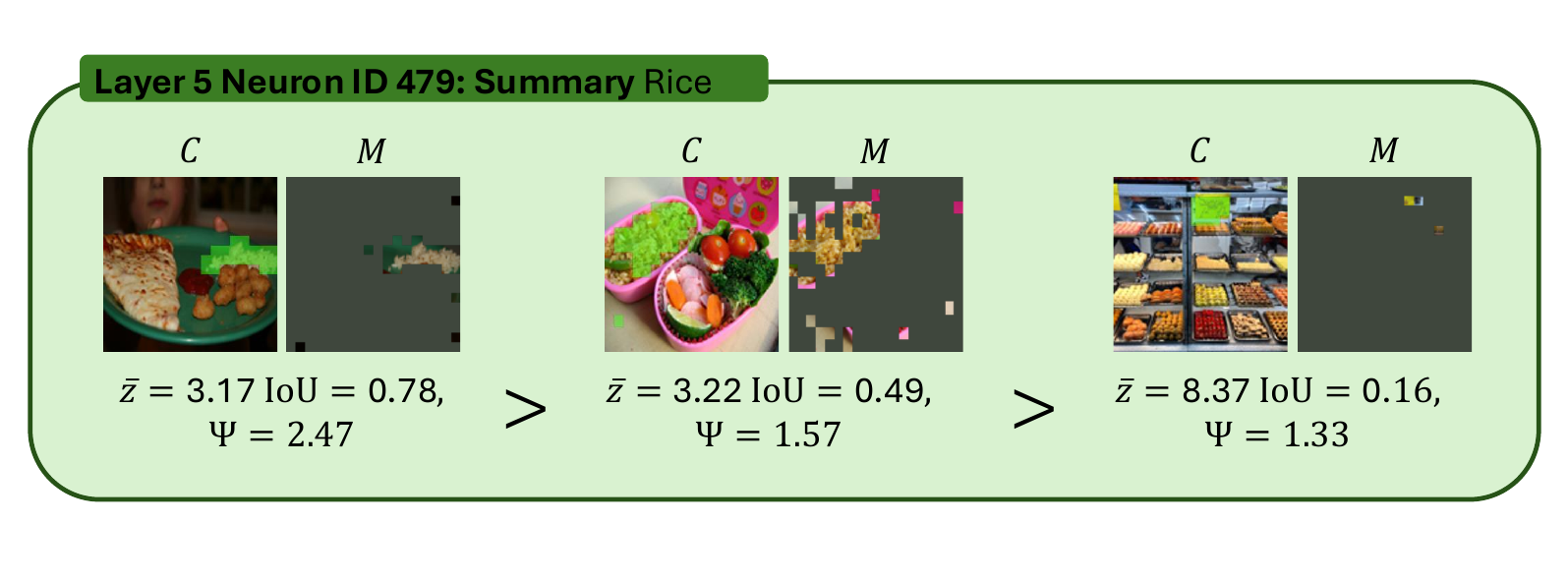}
\caption{
Illustration of activation-semantic alignment.
The cluster \(C\) is highlighted in \textcolor{ForestGreen}{green} in the original image, and the corresponding activation mask \(M\) is shown beside. 
Three examples of vision references show increasing activation intensity (\(\bar{z}\)) but decreasing IoU (from left to right).
The \textbf{leftmost} image serves as the most prototypical representation of the concept $C$ (e.g., \texttt{rice}), exhibiting high conceptual integrity. 
In contrast, the \textbf{middle} image demonstrates concept interference, where the target concept co-occurs with extraneous semantic elements (e.g., \texttt{table}). 
The \textbf{rightmost} image represents a spurious activation triggered by low-level visual noise, such as chromatic similarity between the shelf background and the target \texttt{rice} concept.
}
\label{fig:ref-img-ranking}
\end{figure}

Specifically for vision modality, to bridge the gap between low-level SAE feature activations and high-level semantic concepts, we introduce a Activation-Semantic Alignment mechanism. 
This process filters the reference samples by measuring the structural congruence between a SAE feature's activation footprint and the model’s internal concept clusters.
For each input image, we utilize the previously described clustering (based on attention and spatial information, in \sectionautorefname~\ref{sec:concept_extraction_clustering}) to generate a set of disjoint clusters $\mathcal{C} = \{C_1, C_2, \dots, C_G\}$. 
These clusters serve as the endogenous \textbf{ground truth labels} for latent vision concepts.
For a given SAE feature $j$, let $M_j \in \{0,1\}^{h \times w}$ denote the binary activation mask, where $M_{j,p} = 1$ if the activation at patch $p$ is positive.
We quantify the semantic purity of a SAE feature by calculating the Intersection over Union (IoU) between its mask $M_j$ and each cluster $C_g \in \mathcal{C}$. 
The maximum alignment score is defined as:
\begin{equation}
    \text{IoU}_{j} = \max_{C_g \in \mathcal{C}} \frac{|M_j \cap C_g|}{|M_j \cup C_g|}.
\end{equation}
Rather than relying solely on raw activation magnitude, we rank and select reference samples based on a composite metric $\Psi_j$, defined as the product of the mean activation intensity ($\bar{z}_j$) and the maximum structural alignment ($\text{IoU}_j$):
\begin{equation}
    \Psi_j = \bar{z}_{j} \times \text{IoU}_{j}.
\end{equation}
Only samples with the highest $\Psi_j$ scores are retained for the final interpretation pipeline.
The introduction of the $\Psi_j$ metric offers several practical advantages over vanilla selection based solely on activation values, as illustrated in \figureautorefname~\ref{fig:ref-img-ranking}:
1) \textbf{Suppression of polysemantic noise} Standard selection is susceptible to ``spurious activations'' where a SAE feature might fire strongly but sporadically across unrelated spatial regions. 
By incorporating $\text{IoU}$, we ensure that the SAE feature is consistently dedicated to a single, coherent semantic unit. 
Samples where a SAE feature's activation is scattered across multiple clusters result in a low IoU, effectively filtering out polysemantic noise.
2) \textbf{Endogenous consistency} Since the clusters $\mathcal{C}$ are derived from the VLM's own attention mechanism, they represent the model's internal ``object-level'' logic. 
Aligning SAE features with these clusters ensures that the features we interpret are grounded in the model's own world-view, rather than being arbitrary artifacts of the SAE training process.
3) \textbf{Improved interpretability} By passing only structurally aligned patches to Qwen3-VL, we significantly reduce visual ambiguity. 
The model is presented with a masked image where the remaining content is guaranteed to be a complete, semantically meaningful entity (e.g., a whole eye or a specific texture) rather than a fragmented set of pixels. 
This leads to more accurate and concise linguistic summaries in the final interpretation stage.

\subsection{Concept Interpretation via Modality-aware and Hierarchical Semantic Synthesis}
\label{sec:hierachical_pipeline}


To synthesize the high-level semantic meaning of the learned SAE features, we implement a hierarchical, \textbf{two-stage} interpretation pipeline that unifies representations across modalities. 
In the first stage, a VLM performs localized concept identification, translating masked visual regions into precise linguistic descriptors. 
The second stage utilizes a larger, more capable LLM to perform semantic summarization and measuring consistency, 
aggregating the VLM-generated labels with raw textual samples to verify the SAE feature's conceptual 
alignment. 
This strategic decoupling, delegating fine-grained visual recognition to the VLM while leveraging the LLM for abstract semantic reasoning, is grounded in the automated interpretability framework established by~\citet{bills2023language} and further validated for multi-modal contexts by \citet{zhang2025large}. 
Intuitively, this hierarchical decomposition aligns with the varying complexity of interpretability tasks: the model addresses the relatively straightforward task of individual concept recognition (i.e., ``\textit{What is present in this image?}'') before tackling the more cognitively demanding task of abstract semantic synthesis (i.e., ``\textit{What commonalities link these disparate cases?}'').
This approach ensures that the interpretation process is both scalable and semantically rigorous, bypassing the ambiguity of raw visual activations by projecting them into a unified linguistic space for final synthesis.

\paragraph{Concept identification}
With the reference samples collected, the \textbf{first stage} of our interpretability pipeline involves translating the visual features activated by each SAE feature into a format compatible with automated reasoning.
Unlike the language modality, where concepts are inherently represented as natural language tokens, visual concepts are represented as masking within an image. 
To facilitate the use of LLMs for high-level semantic synthesis, we must first convert these masked visual regions into precise linguistic descriptors.
We utilize \texttt{Qwen3-VL-8B-Instruct} as a high-level visual annotator. 
For each visual sample (with an original image and a corresponding mask associated with one SAE feature), it is prompted to describe which visual feature or concept is indicated within the masked regions and provides a brief explanation that encapsulates its behavior. 
The detailed prompt is given in Prompt~\ref{prompt:image_identification}.
By leveraging the advanced fine-grained visual recognition capabilities of \texttt{Qwen3-VL}, we obtain a set of descriptive ``visual labels'' for each SAE feature, which serves as a textual proxy for its visual preferences.

\paragraph{Dilation} 
To ensure the VLM can effectively contextualize the features of interest, we apply a dilation operation to both the visual and textual masks. 
Specifically, we expand the mask boundaries by one unit (i.e., extending by one patch in the image grid and one token in the text sequence), thereby providing the model with localized peripheral information necessary for accurate concept identification.

\paragraph{Summarizing references via LLM}
Once visual concepts are translated into natural language, the vision modality can be processed using the same unified pipeline as the language modality in the \textbf{second stage}.
We perform a holistic semantic synthesis using \texttt{Qwen3-30B-Instruct}. 
This model acts as the central ``interpreter'' by consuming two sources of information for each SAE feature: 
1) The visual concept identifications generated by \texttt{Qwen3-VL}; 
2) The raw masked text references that triggered this SAE feature.
\texttt{Qwen3-30B} is tasked with generating a comprehensive explanation of the SAE feature’s function and determining its modality. 
Specifically, the model first performs intra-modality summarization to identify the core themes within the vision and language samples independently using Prompt~\ref{prompt:summarization}. 
Subsequently, it performs a cross-modality comparison between these two summaries to determine their semantic consistency using Prompt~\ref{prompt:consistency}.

\section{Experiments}

The experimental parts answer two core research questions: (1) \textit{whether the proposed structural regularizers enable S$^2$AE to learn grounded and coherent visual concepts without degrading reconstruction performance}, and (2) \textit{whether the proposed hierarchical interpretation pipeline provides more reliable SAE feature summaries than direct VLM summarization}.
Specifically, we evaluate whether the learned vision concepts are both semantically distinct and group-consistent across vision patches. 
We deploy our proposed S$^2$AE architecture on the residual streams of \texttt{Qwen2.5-VL-7B-Instruct} (e.g., Layers 5, 10, 15, 20). 
Given the high dimensionality of the residual blocks, we train the TopK-SAE with an expansion factor of $R=32$ and \(K=256\). 
For the details of the implementation and training settings, please refer to Appendix~\ref{sec:impl_details}. 

\subsection{Results}


\begin{figure}[ht]
\centering
\includegraphics[width=\textwidth]{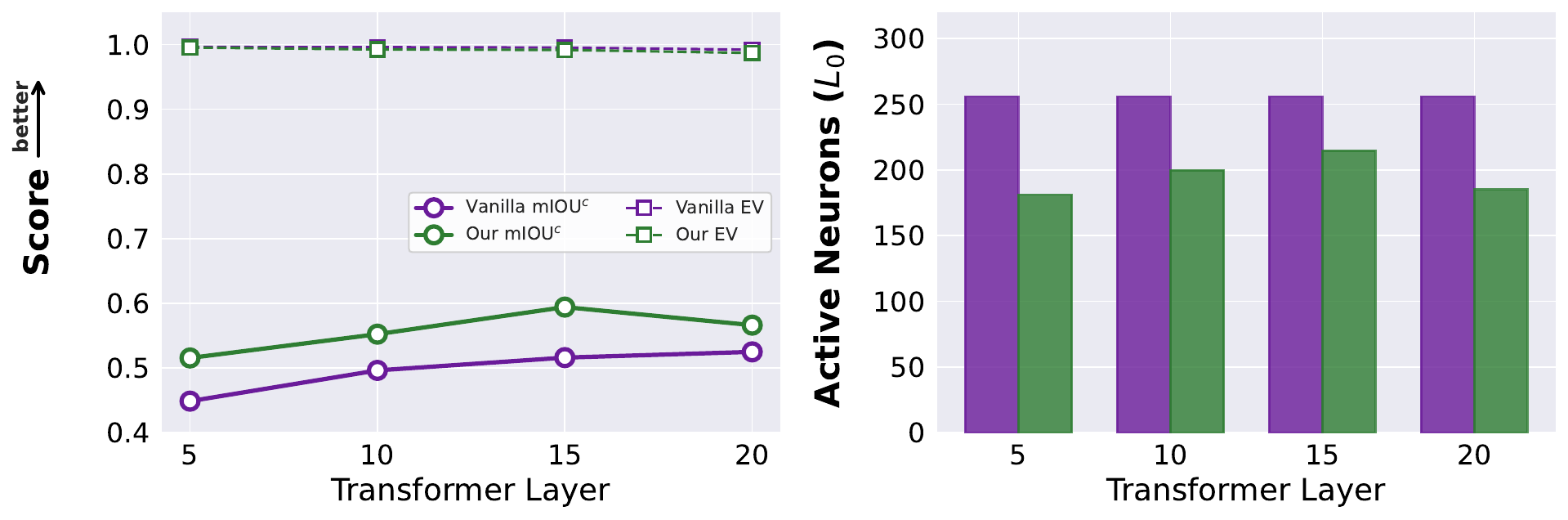}
\caption{
Quantitative comparison of SAE performance and sparsity across Transformer layers.
\textbf{Left:} The trade-off between reconstruction fidelity (Explained Variance, EV, dashed lines) and semantic alignment (\(\text{mIOU}^c\), solid lines). 
Our method significantly improves semantic alignment with vision labels while maintaining a near-identical reconstruction score ($>0.99$) compared to the vanilla baseline. 
\textbf{Right:} Sparsity cost measured by the number of active SAE features ($\ell_0$ norm). 
\textbf{Takeaway:} Our approach achieves superior isolation of concepts using significantly fewer active SAE features across all evaluated layers, demonstrating higher representational efficiency.
}
\label{fig:main}
\end{figure}

To qualitatively evaluate the effectiveness of our proposed structural regularizers, we randomly collect 5000 images from the reference dataset and calculate the average maximum alignment score (\(\text{mIoU}^c\)) over all activated SAE features. 
That is, \(\text{mIoU}^c=\text{Mean}_{j\in N^+}\left(\text{IoU}_j\right)\), where \(N^+\) is a collection of activated SAE features.  
Moreover, we report the explained variance (EV) to evaluate the reconstruction fidelity and the average number of active SAE features (\(L_0\) norm) to evaluate the sparsity cost. 

\paragraph{S$^2$AE improves visual concept alignment while preserving reconstruction fidelity and increasing sparsity efficiency}
According to the statistical results shown in \figureautorefname~\ref{fig:main}, 
S$^2$AE demonstrates a consistent and robust improvement in the alignment between SAE feature activation patches and ground-truth clusters across the model hierarchy (from layers 5 to 20). 
In addition, a layer-wise comparison reveals that the performance gains are particularly pronounced in the shallow or middle layers of the model hierarchy, where the transition from raw visual signals to structured semantic concepts is most evident.
Despite the gains in concept isolation, the explained variance for S$^2$AE remains remarkably stable and high (above 0.99). 
This indicates that S$^2$AE successfully disentangles semantic concepts without sacrificing the reconstruction quality of the vision patches.
This enhanced structural correspondence significantly strengthens the SAE’s capacity for accurate visual concept identification.
The right panel demonstrates that S$^2$AE maintains a significantly lower $L_0$ norm across all layers. 
While the vanilla SAE is fixed at a budget of 256 active SAE features, our approach utilizes between 180 and 215 SAE features, achieving a more efficient and sparse representation of the input data without noisy SAE activation.


\subsection{Modality Summary Statistics}


\begin{table}[htbp]
  \centering
  \caption{Modality summarization based on interpretation.}
  \label{table:modality-summary}
  \begin{tabular}{r|cc|cc|cc|cc}
    \toprule
    & \multicolumn{2}{c|}{\textbf{Layer 5}} & \multicolumn{2}{c|}{\textbf{Layer 10}} & \multicolumn{2}{c|}{\textbf{Layer 15}} & \multicolumn{2}{c}{\textbf{Layer 20}} \\
    & Vanilla & S$^2$AE & Vanilla & S$^2$AE & Vanilla & S$^2$AE & Vanilla & S$^2$AE \\
    \midrule
    \textbf{Identification rate} & 0.988 & 0.992 & 0.987 & 0.978 & 0.981 & 0.971 & 0.966 & 0.971 \\
    \midrule
    \textbf{Multi-modal rate}    & 0.772 & 0.721 & 0.770 & 0.482 & 0.681 & 0.404 & 0.656 & 0.549 \\
    \textbf{Consistency rate  \(\uparrow\)}   & 0.294 & \textbf{0.337} & 0.327 & \textbf{0.366} & 0.322 & \textbf{0.348} & 0.308 & \textbf{0.319} \\
    \textbf{Vision monosemanticity  \(\uparrow\)} & 0.510 & \textbf{0.546} & 0.501 & \textbf{0.534} & 0.492 & \textbf{0.518} & 0.490 & \textbf{0.501} \\
    \textbf{Language monosemanticity  \(\uparrow\)} & 0.872 & \textbf{0.896} & 0.868 & \textbf{0.903} & 0.869 & \textbf{0.881} & 0.862 & \textbf{0.875} \\
    \midrule
    \textbf{Vision rate}         & 0.178 & 0.263 & 0.185 & 0.502 & 0.272 & 0.581 & 0.272 & 0.423 \\
    \textbf{Vision monosemanticity  \(\uparrow\)} & 0.495 & \textbf{0.517} & \textbf{0.487} & 0.445 & \textbf{0.479} & 0.441 & \textbf{0.470} & 0.454 \\
    \midrule
    \textbf{Language rate}       & 0.050  & 0.016 & 0.045 & 0.016  & 0.047  & 0.015  & 0.072  & 0.028 \\
    \textbf{Language monosemanticity  \(\uparrow\)} & \textbf{0.873} & 0.858 & 0.863 & \textbf{0.888} & 0.864 & \textbf{0.914} & 0.870 & \textbf{0.905} \\
    \bottomrule
  \end{tabular}
\end{table}

In this subsection, we summarize the statistics of S$^2$AE and vanilla SAE features w.r.t. modalities in \tableautorefname~\ref{table:modality-summary}, respectively.
Specifically, \textbf{Identification rate} refers to the percentage of valid SAE features that contain either a vision summary or a language summary or both. 
An invalid SAE feature means LLM can not give a clear summary on both vision references and language references. 
Among these valid SAE features, \textbf{Multi-modal rate, Vision rate, and Language rate} refer to the percentages of SAE features in specific modalities, respectively. 
\textbf{Consistency rate} refers to the percentage of multi-modal SAE features whose vision and language summaries have a consistent score \(> 0.0\). 
\textbf{Vision monosemanticity} and \textbf{Language monosemanticity} refer to the MS score~\citep{pach2026sparse} w.r.t. vision and language reference samples, respectively. 
We report the MS scores for different modalities and use SigLip2~\citep{tschannen2025siglip} to extract vision and language embeddings on active tokens, as it is a recent vision-language encoder optimized for cross-modality tasks, making it a suitable external semantic space for comparing visual and textual reference samples.

\paragraph{S$^2$AE yields more semantically consistent cross-modal features}
The Identification rate remains exceptionally high across all layers ($>97\%$), suggesting that the vast majority of active SAE features capture meaningful concepts that can be mapped to vision or language summaries. 
There is a slight downward trend as the layers deepen, potentially reflecting the increased abstraction and complexity of concepts in higher layers of the Transformer.
A striking observation is the disparity between unimodal SAE features. 
The Vision rate is significantly higher than the Language rate across all layers, peaking at $58.10\%$ in Layer 15. 
This indicates that the SAE primarily decomposes the latent space into visual primitives or grounded visual-language concepts rather than isolated language concepts.
Despite the high multi-modal rate, S$^2$AE consistently achieves a higher consistency rate than the vanilla SAE across all layers (peaking gain at $4.3\%$ in Layer 5), demonstrating that the application of group and exclusive sparsity effectively optimizes vision-based SAE feature performance by reducing noise and polysemy while enhancing cross-modal alignment between vision and language.

\paragraph{S$^2$AE improves multi-modal feature monosemanticity beyond the directly regularized vision side}
\tableautorefname~\ref{table:modality-summary} shows that for multi-modal SAE features, S$^2$AE consistently improves not only vision monosemanticity but also language monosemanticity across all evaluated layers, although the structural regularizers are applied only to visual patch groups. 
For example, language MS increases from $0.872\!\to\!0.896$ at Layer 5, accompanied by consistent gains in vision-language consistency rate. 
This suggests that visual structure can act as an organizing prior for the shared multi-modal SAE feature space. 
By suppressing fragmented or background-driven visual activations, S$^2$AE helps each multi-modal feature acquire a cleaner semantic meaning, which in turn makes its language-side activating contexts more coherent. 
We emphasize that this effect should be interpreted as an emergent cross-modal alignment effect through shared SAE features, rather than direct regularization of language tokens.


\subsection{Concept Analysis with Word Cloud}

\begin{figure}[ht]
\centering
\begin{subfigure}[b]{0.23\textwidth}
    \centering
    \includegraphics[width=\textwidth]{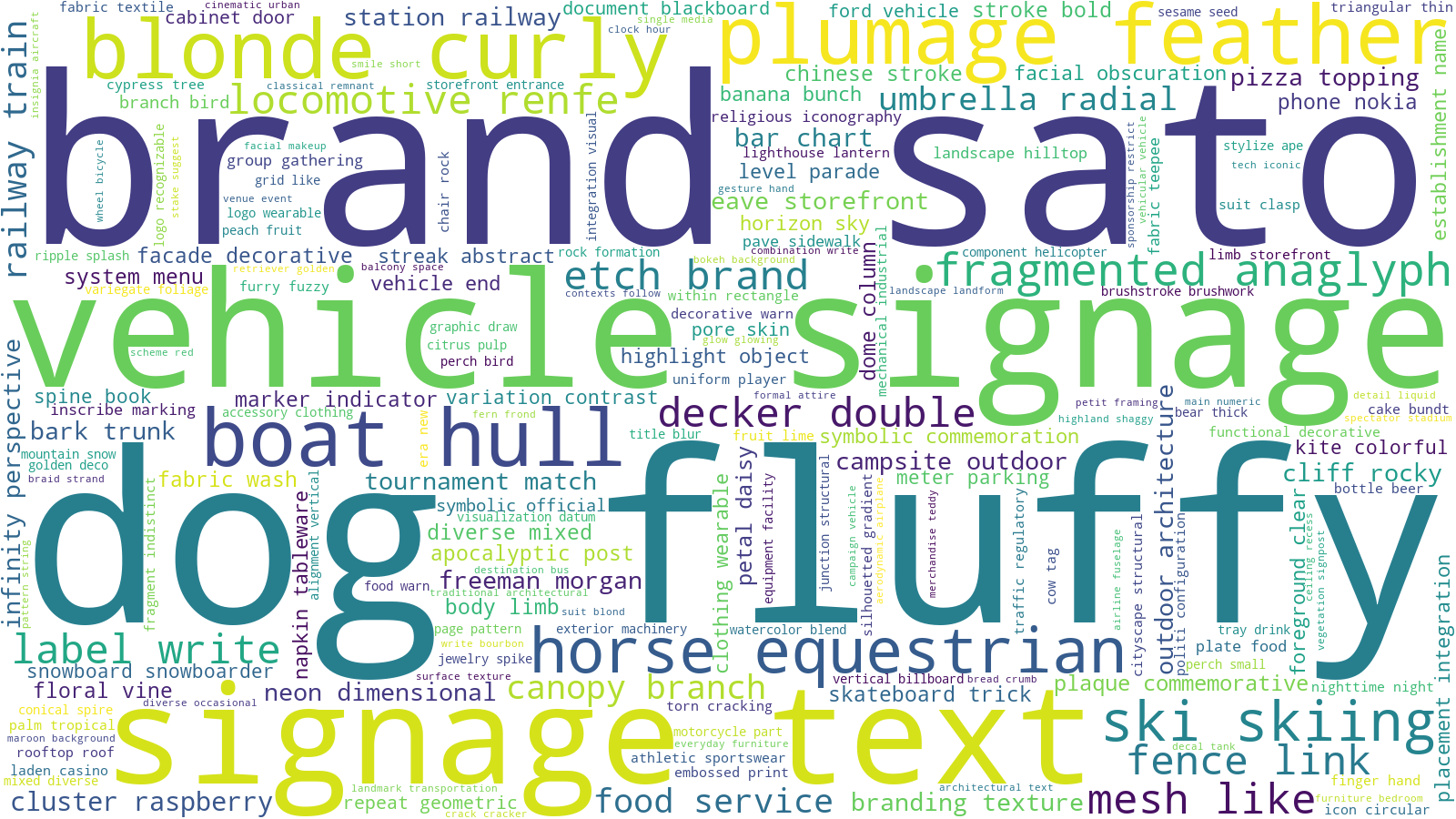}
    \caption{Vision (\(l_5\))}
    \label{fig:wc:5:img}
\end{subfigure}
\hfill
\begin{subfigure}[b]{0.23\textwidth}
    \centering
    \includegraphics[width=\textwidth]{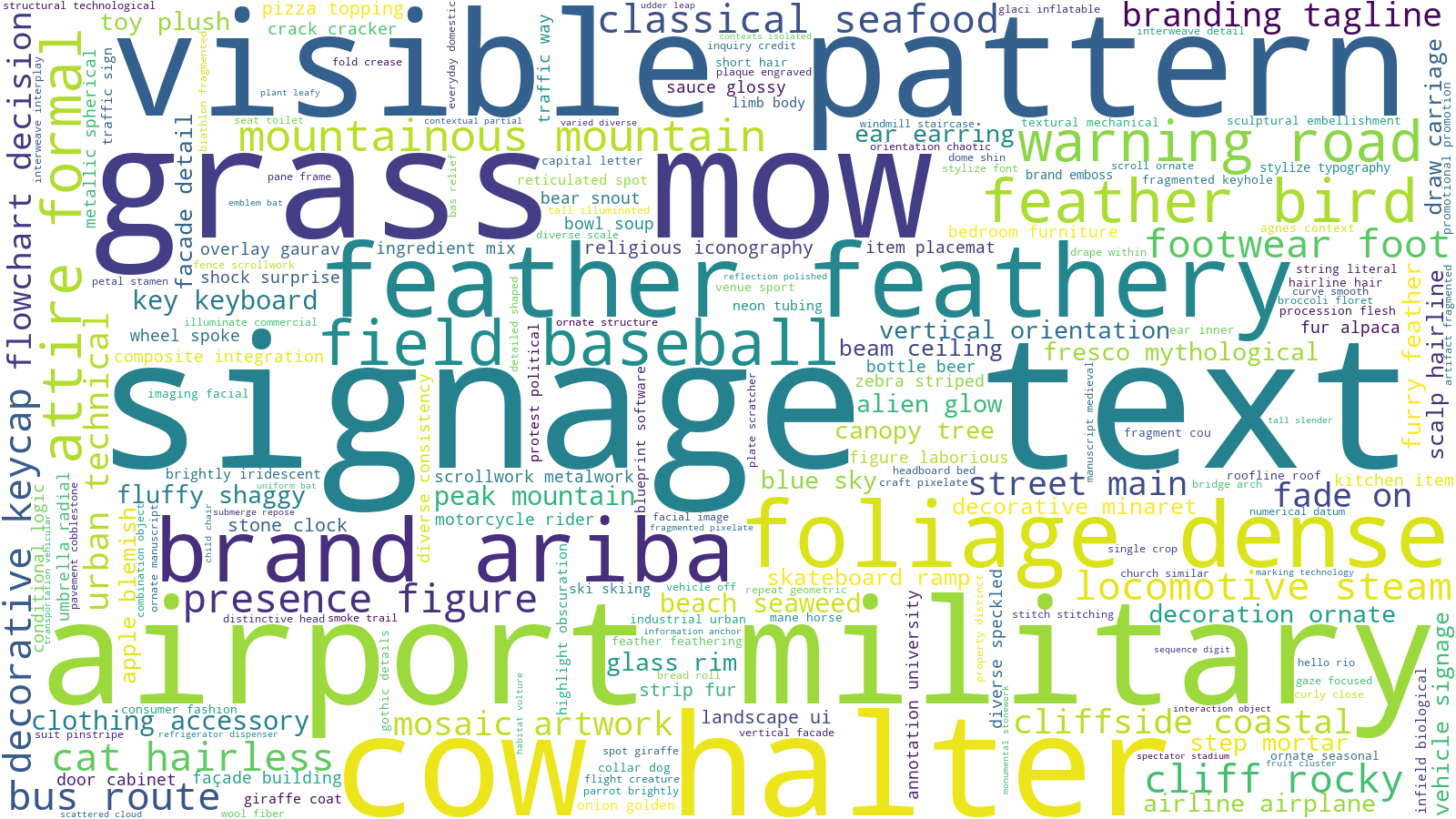}
    \caption{Vision (\(l_{10}\))}
    \label{fig:wc:10:img}
\end{subfigure}
\hfill
\begin{subfigure}[b]{0.23\textwidth}
    \centering
    \includegraphics[width=\textwidth]{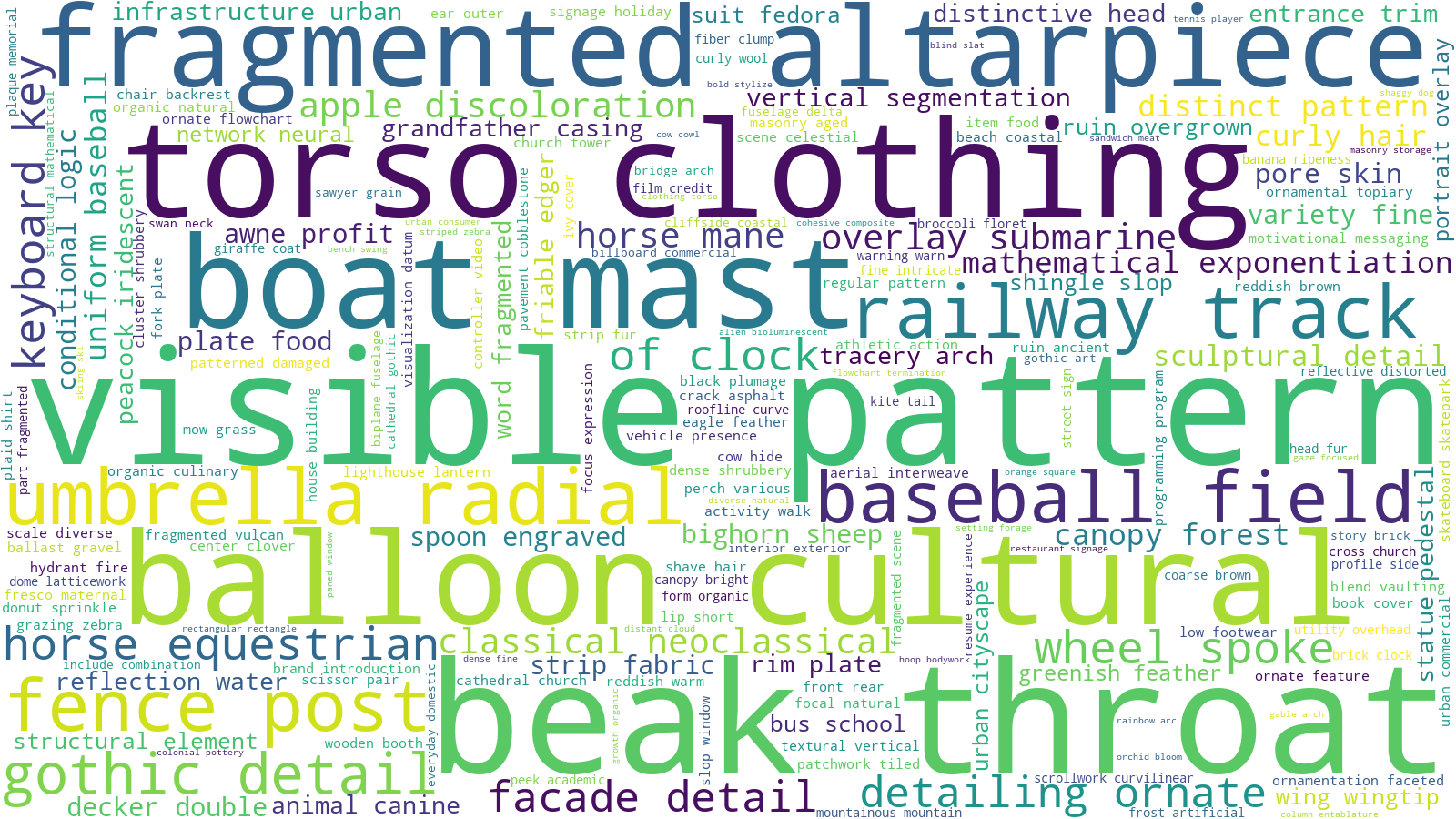}
    \caption{Vision (\(l_{15}\))}
    \label{fig:wc:15:img}
\end{subfigure}
\hfill
\begin{subfigure}[b]{0.23\textwidth}
    \centering
    \includegraphics[width=\textwidth]{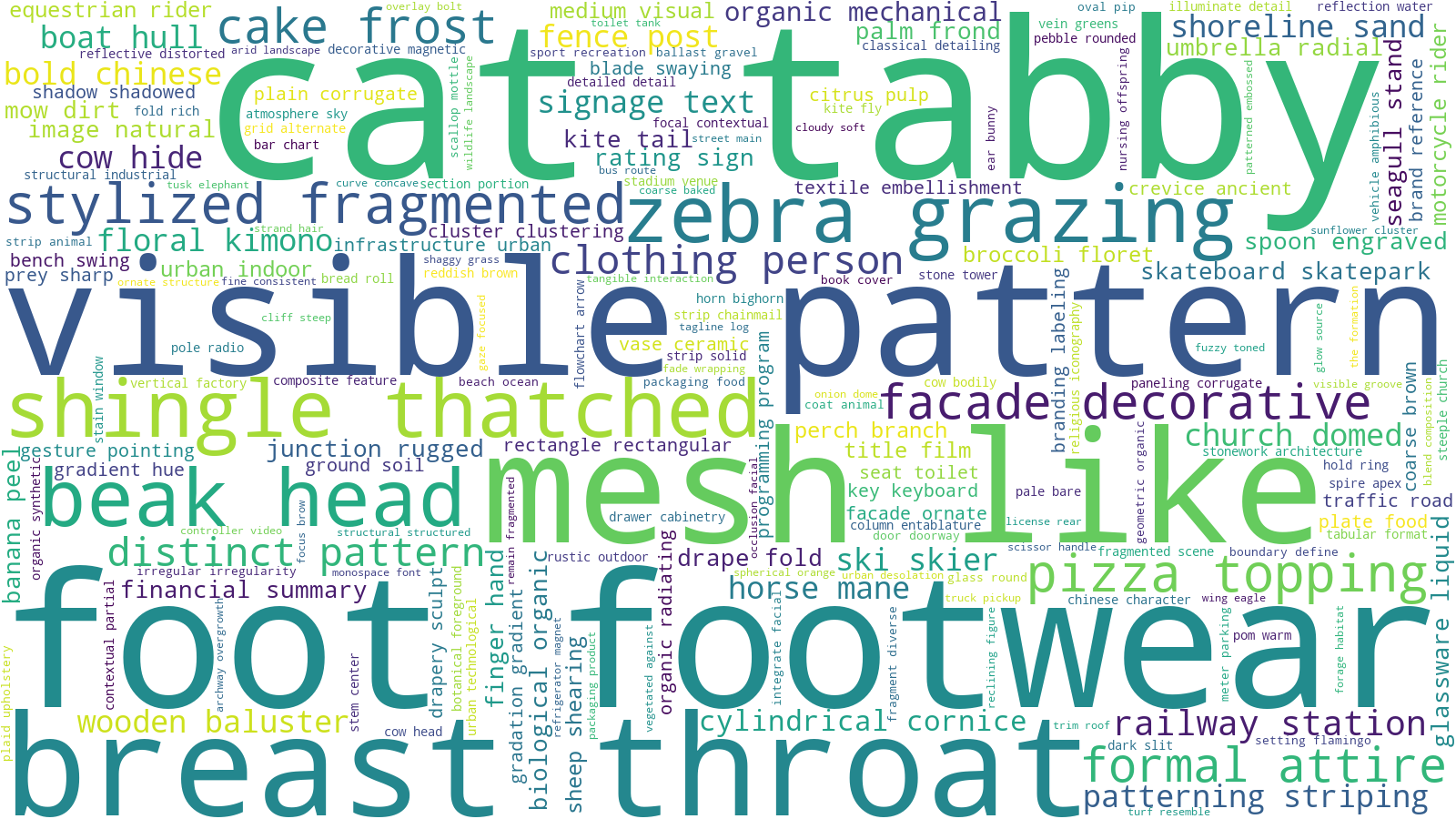}
    \caption{Vision (\(l_{20}\))}
    \label{fig:wc:20:img}
\end{subfigure}
\hfill
\begin{subfigure}[b]{0.23\textwidth}
    \centering
    \includegraphics[width=\textwidth]{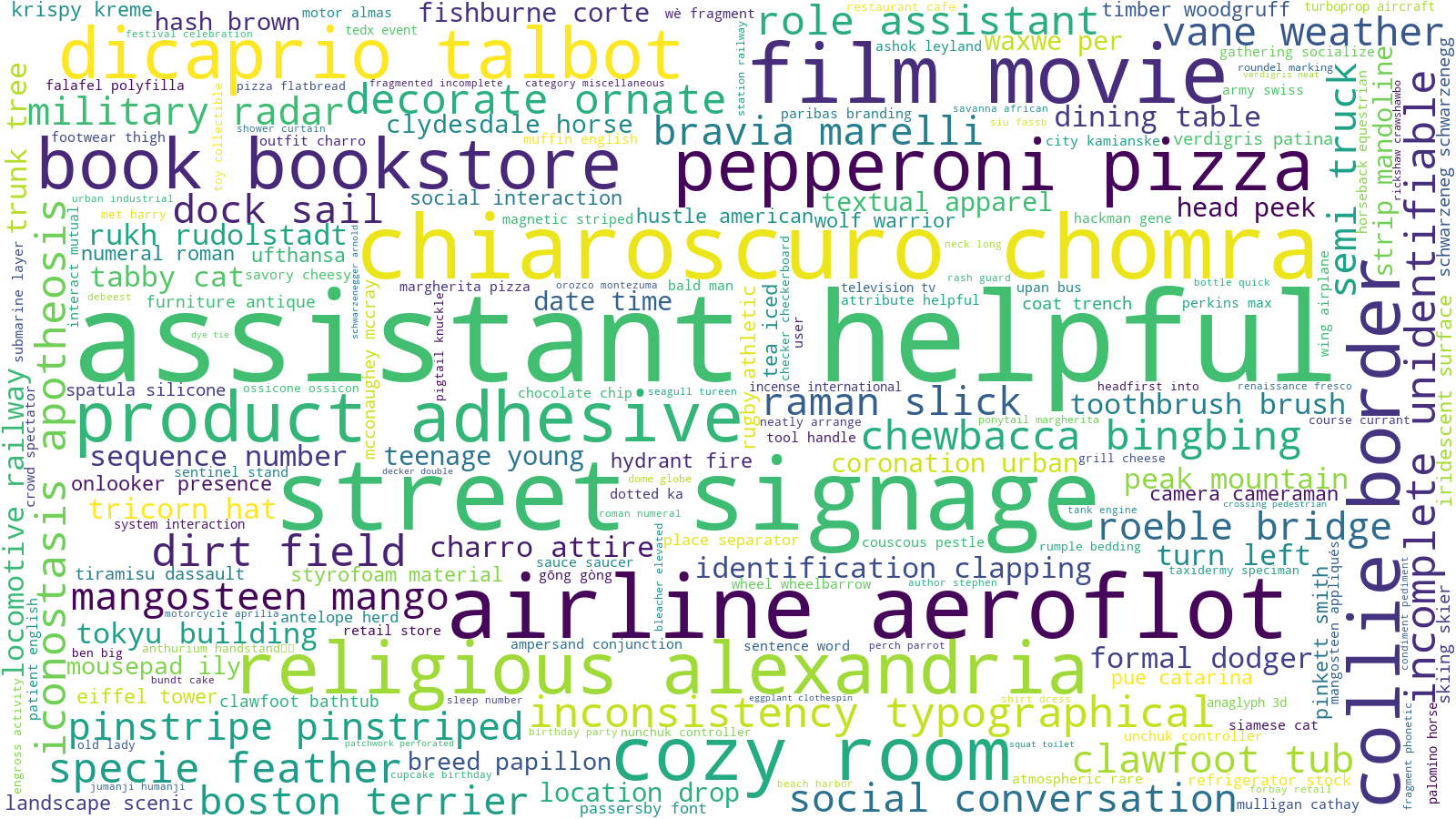}
    \caption{Language (\(l_5\))}
    \label{fig:wc:5:txt}
\end{subfigure}
\hfill
\begin{subfigure}[b]{0.23\textwidth}
    \centering
    \includegraphics[width=\textwidth]{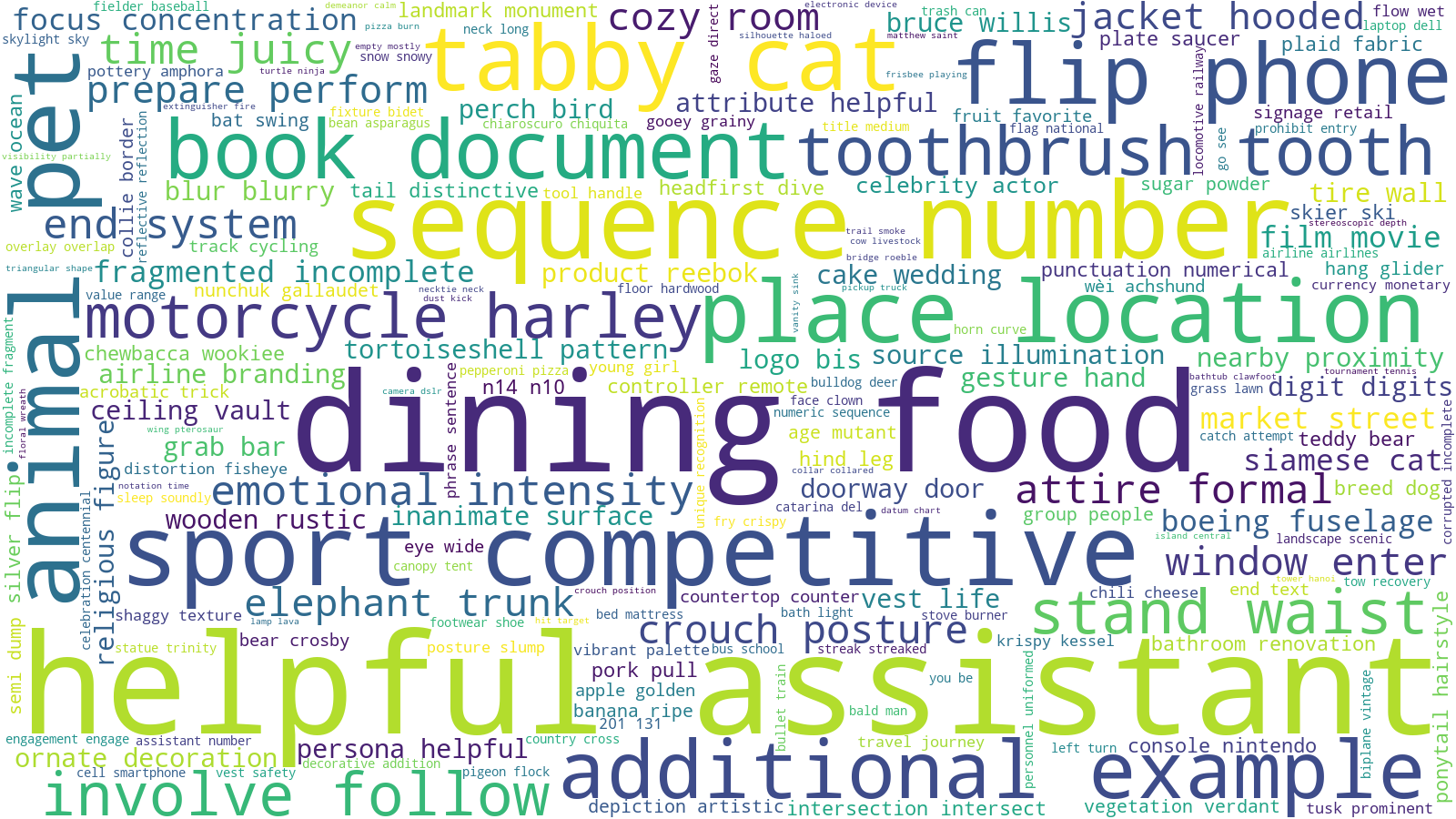}
    \caption{Language (\(l_{10}\))}
    \label{fig:wc:10:txt}
\end{subfigure}
\hfill
\begin{subfigure}[b]{0.23\textwidth}
    \centering
    \includegraphics[width=\textwidth]{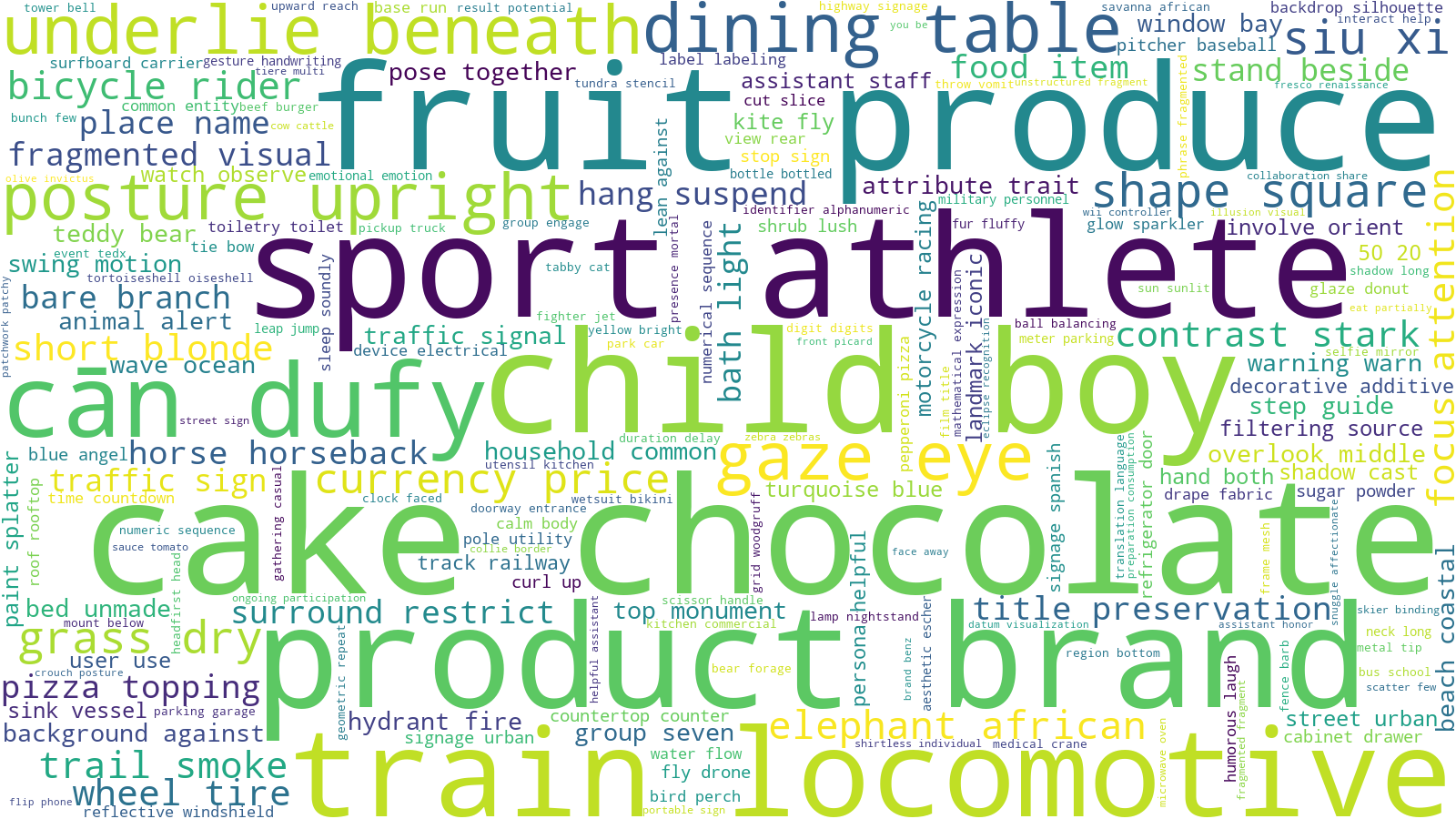}
    \caption{Language (\(l_{15}\))}
    \label{fig:wc:15:txt}
\end{subfigure}
\hfill
\begin{subfigure}[b]{0.23\textwidth}
    \centering
    \includegraphics[width=\textwidth]{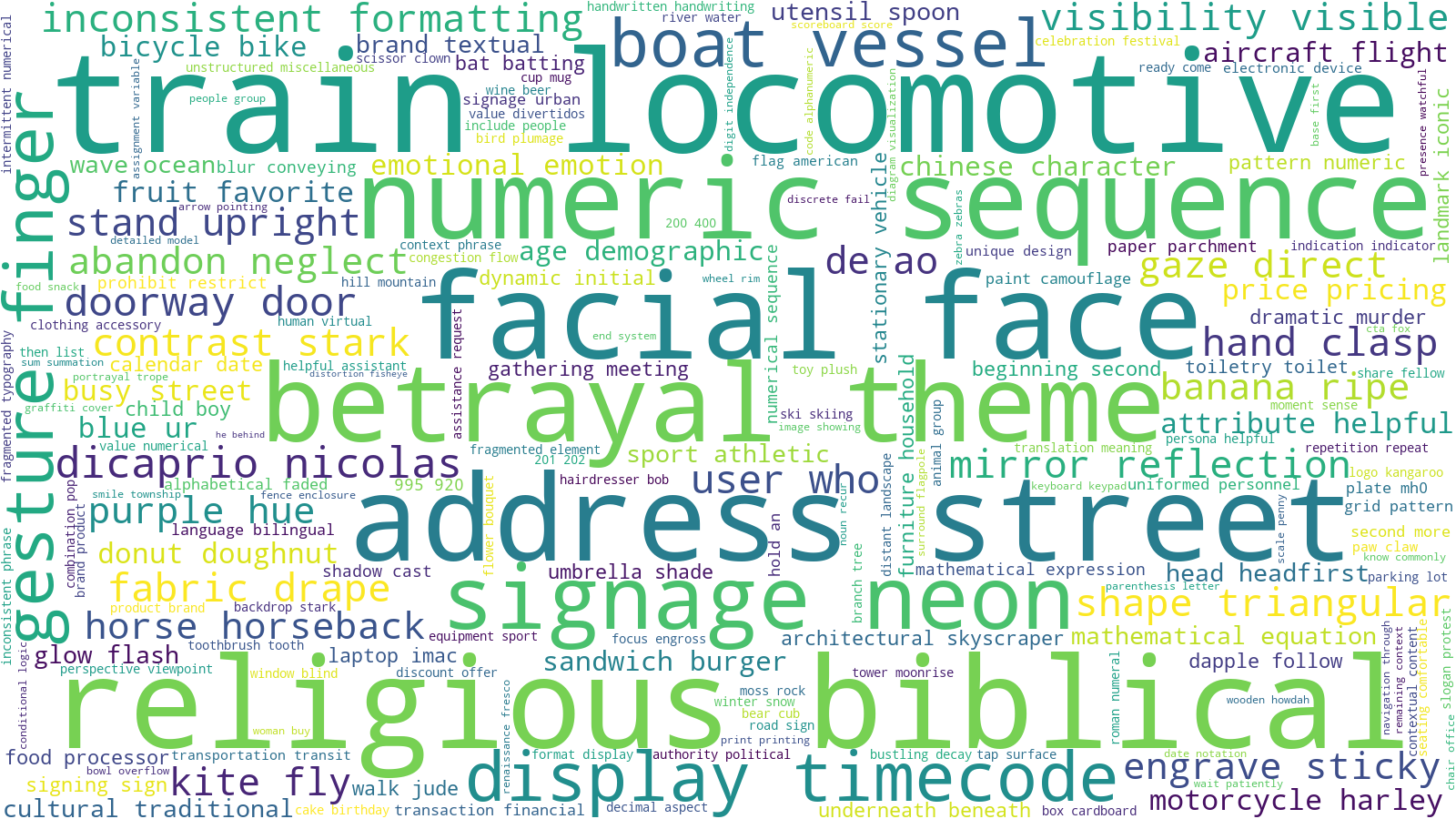}
    \caption{Language (\(l_{20}\))}
    \label{fig:wc:20:txt}
\end{subfigure}
\caption{
Evolution of Semantic Distributions in Vision and Language modalities. 
These word clouds visualize the thematic shift of concepts across four representative layers ($l_5, l_{10}, l_{15}, l_{20}$). 
\textbf{Takeaway:} the network undergoes a systematic transition from modality-specific sensory perception to a converged, high-level conceptual space where visual morphology and linguistic narratives align through shared symbolic abstractions.
}
\label{fig:word-cloud}
\end{figure}

To facilitate a comparative analysis of visual and language features, we employ an automated interpretation pipeline that maps vision-based activations into the natural language domain. 
This allows for a unified characterization of features across modalities.
To rigorously quantify the distributional shifts of concepts across different SAE layers, we implemented a multi-stage text processing pipeline.
First, we perform linguistic normalization with \texttt{en\_core\_web\_sm} model in the spaCy software~\citep{honnibal2020spacy} to mitigate noise from morphological variations (e.g., tense, plurality). 
Each summary generated from the SAE features is lemmatized, ensuring that semantically identical concepts are mapped to a single canonical form.
However, the high cardinality of unique lemmas posed a challenge for direct interpretation. 
To address this high-dimensional sparsity, we employ BERTopic~\citep{grootendorst2022bertopic} to group semantically related terms into cohesive clusters, including the steps as follows:
1) First, we utilize the \texttt{all-MiniLM-L6-v2} SentenceTransformer~\citep{reimers-2019-sentence-bert} to generate dense semantic embeddings.
2) Next, these embeddings undergo dimensionality reduction (i.e., UMAP~\citep{mcinnes2018umap}) and hierarchical clustering (i.e., HDBSCAN~\citep{mcinnes2017hdbscan}) to distill the raw concepts into distinct, high-level ``topic words''.
3) Finally, these ``topic words'' are visualized as word clouds in \figureautorefname~\ref{fig:word-cloud}.

Across layers, the word clouds reveal a clear shift in the granularity and abstraction of SAE concepts, with early layers focusing on concrete perceptual cues and later layers moving toward structured, symbolic semantics. 
At Layer 5, vision concepts are dominated by low-level perceptual and object-level attributes such as \texttt{fluffy} and \texttt{vehicle}, while language concepts already show functional descriptors such as \texttt{assistant} and \texttt{helpful}. 
At Layer 10, concepts become more contextual. Vision features capture scene-level environments such as \texttt{airport} and \texttt{stadium}, whereas language features emphasize relational or behavioral notions such as \texttt{competitive} and \texttt{intensity}. 
At Layer 15, vision concepts further specialize into fine-grained morphology and style, including \texttt{beak}, \texttt{torso}, \texttt{gothic}, and \texttt{ornate}, while language concepts concentrate on human-centric and socio-economic categories such as \texttt{product}, \texttt{brand}, \texttt{athlete}, and \texttt{child}. 
At Layer 20, the concepts become the most abstract. Vision features converge to general morphological primitives such as \texttt{foot}, \texttt{head}, and \texttt{mesh}, while language features shift toward thematic and symbolic concepts such as \texttt{theme}, \texttt{betrayal}, and \texttt{biblical}.

\subsection{Case Studies}

\begin{figure}[ht]
\centering
\includegraphics[width=\textwidth]{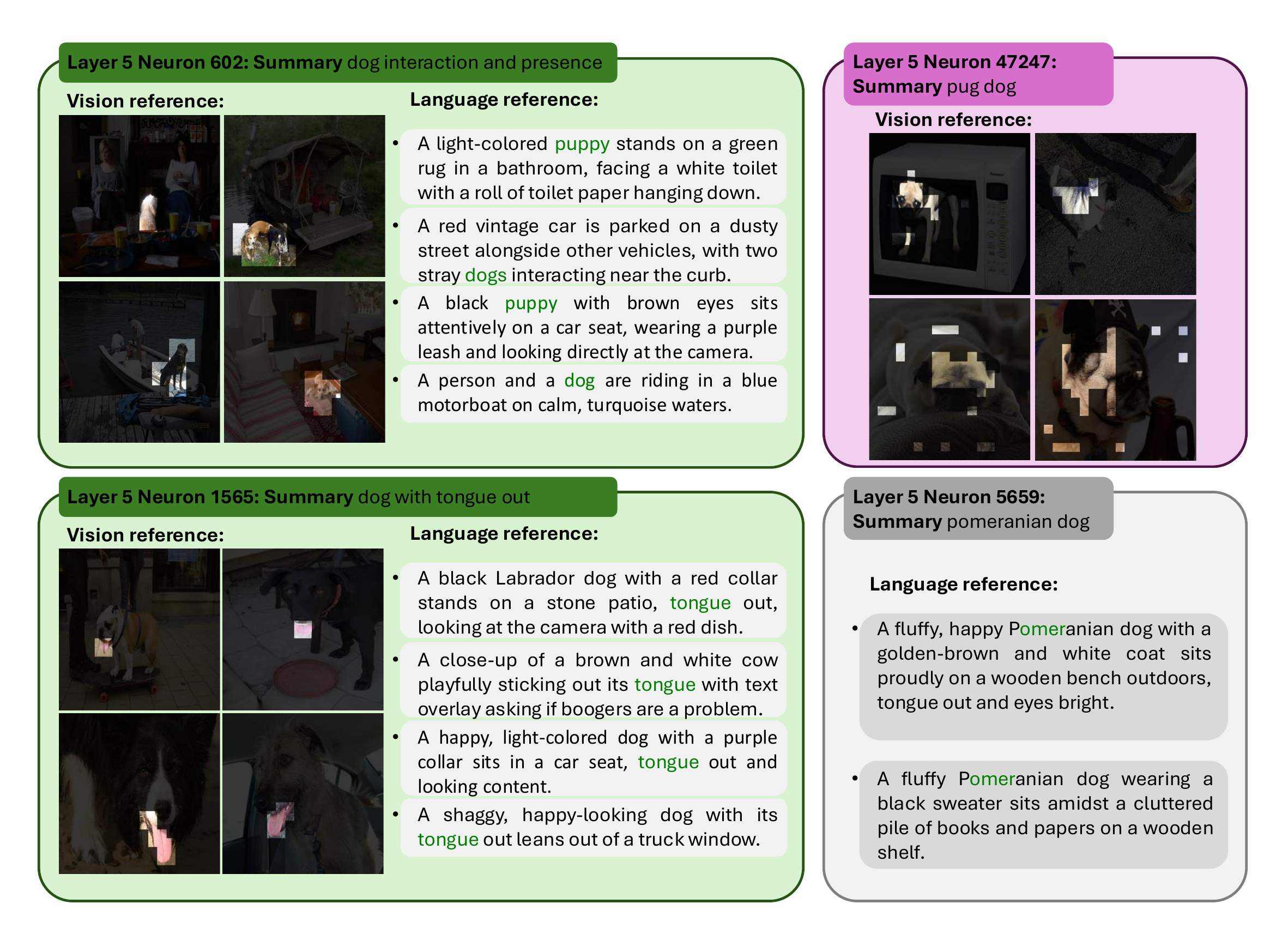}
\caption{
Examples of dog-related SAE features. 
Color indicates the modality of the SAE feature we present. 
Two \textcolor{ForestGreen}{multi-modal} SAE features indicate \texttt{dog} and \texttt{dog tongue} concepts, respectively. 
A \textcolor{purple}{vision} SAE feature indicates a specific dog species: \texttt{pug dog}. 
A \textcolor{gray}{language} SAE feature indicates another species: \texttt{pomeranian dog}.
Vision and language concepts are highlighted with a patch-level mask and in \textcolor{ForestGreen}{green}, respectively.
}
\label{fig:case-study-dog}
\end{figure}

This section selects a set of dog-related SAE features as a presentation of semantic grounding to further elucidate the interpretability of the learned concepts. 
In \figureautorefname~\ref{fig:case-study-dog}, these examples demonstrate how the SAE captures specific semantic nuances across different modalities and activation patterns.
SAE features such as Neuron 602 and Neuron 1565 exhibit robust cross-modal alignment. 
Neuron 602 (\texttt{dog interaction and presence}) is activated by diverse visual scenes of dogs in various environments, ranging from boats to living rooms, paired with linguistic descriptions of \texttt{puppy} and \texttt{dog}. 
Similarly, Neuron 1565 focuses on a specific behavioral trait (\texttt{dog with tongue out}), capturing the precise visual feature of an extended tongue across different dogs and their corresponding textual mentions.
Interestingly, some SAE features show specialized sensitivity within a single modality. 
Neuron 47247
\footnote{This is a S$^2$AE feature with vision and language concepts: \texttt{pug dog} and \texttt{``umpire'' word}, respectively.} 
(\texttt{pug dog}) demonstrates high visual specificity for the unique facial structure of pugs. 
In contrast, Neuron 5659 
\footnote{This is a S$^2$AE feature with vision and language concepts: \texttt{urban skyline with text elements} and \texttt{pomeranian dog}, respectively.}
(\texttt{pomeranian dog}) shows strong linguistic grounding, as it is triggered by complex descriptive sentences detailing the Pomeranian’s fluffy coat and bright eyes, even when the visual cues are varied.
These examples highlight the SAE's ability to decouple broad categories (e.g., \texttt{dog}) into fine-grained sub-concepts based on behavior (\texttt{tongue out}) and breed-specific morphology (\texttt{pug}, \texttt{pomeranian}).










\subsection{Improved Vision Representation}
\label{sec:improved_vision_repr}

\begin{figure}[ht]
\centering
\includegraphics[width=\textwidth]{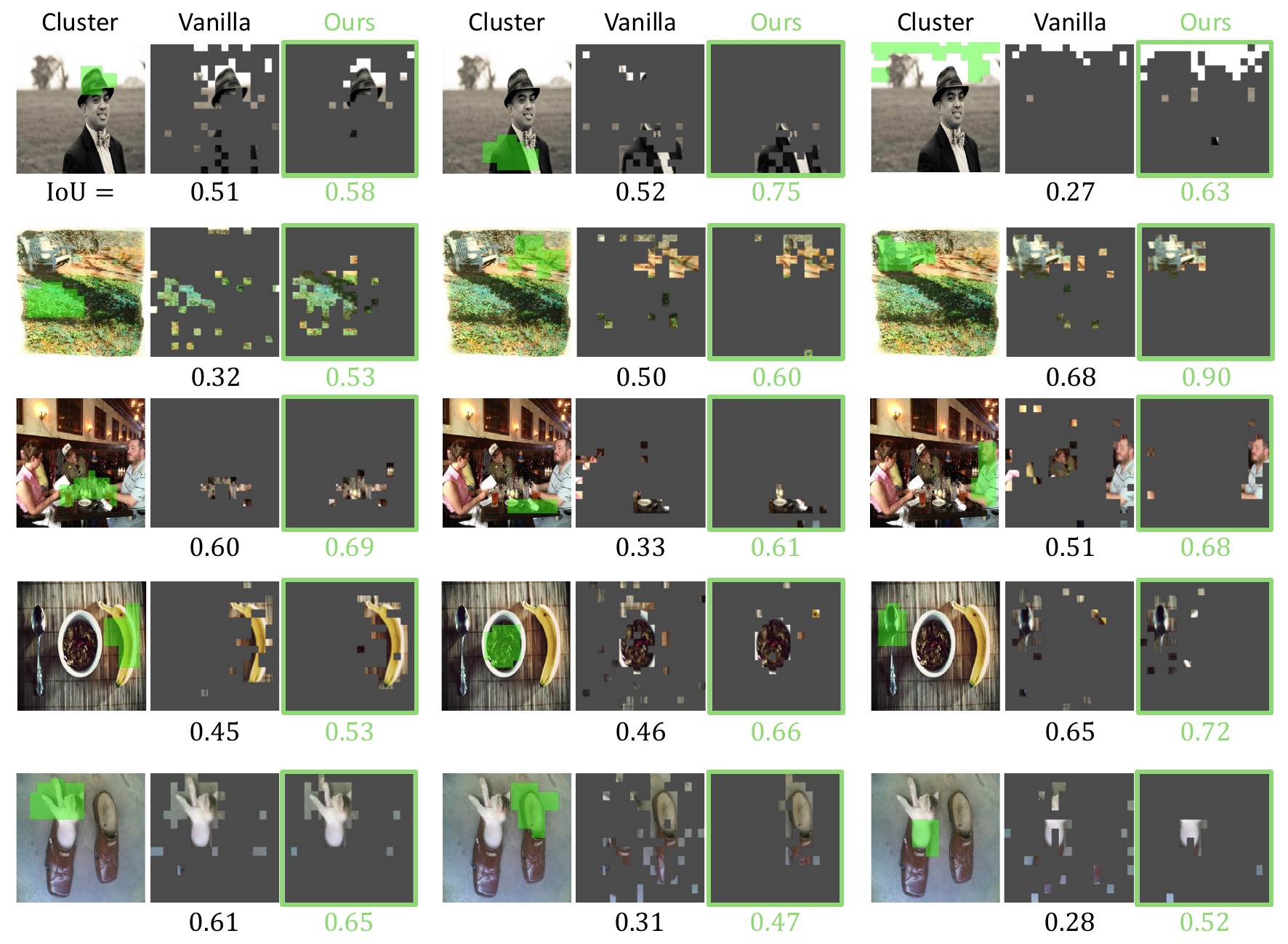}
\caption{
Visualization of SAE activation masks across different clusters. 
SAE is trained on \texttt{Qwen2.5-VL-7B-Instruct}'s layer 5.  
Each row presents an image alongside three distinct cluster cases to compare SAE activation mask performance. 
}
\label{fig:cluster-mask}
\end{figure}

We visualize and compare the activation masks of neurons from both the vanilla SAE and our regularized SAE learned on VLM's layer 5 in \figureautorefname~\ref{fig:cluster-mask}. 
For a given semantic cluster, we identify the neuron that yields the maximum IoU between its activation footprint and the cluster's spatial region. 
This selection ensures that we are comparing the most representative neuron for each specific concept.
Our observations highlight three significant improvements. 

\paragraph{Suppression of polysemantic noise} 
First, our method effectively suppresses polysemantic noise and spurious activations. 
As shown in the first case of the first image, where the target concept is \texttt{hat}, S$^2$AE precisely isolates the hat region. 
In contrast, the vanilla SAE erroneously activates additional patches on the subject's black suit, likely due to chromatic similarity. 
Similarly, in the final case of the second image (concept: \texttt{chair}), our method maintains a focused activation on the chair itself, while the vanilla SAE's activation leaks into surrounding, semantically unrelated areas.

\paragraph{Improvement of concept completeness}
Second, our approach resolves the issue of incomplete or fragmented concept activation. 
In the first case of the third image, the target concept involves \texttt{glasses} on a table. 
The vanilla SAE fails to activate the glass located on the further table, whereas our method successfully captures all relevant instances. 
A similar improvement is observed in the first case of the fourth image (concept: \texttt{banana}), where our SAE provides a much more complete and coherent activation mask compared to the sparse output of the vanilla model.

\paragraph{More fine-grained semantic concepts}
Third, our method captures more fine-grained semantic concepts. 
In the last image, while the vanilla SAE only identifies the \texttt{cat} as a single holistic concept, our model can distinguish between the broader \texttt{cat} concept and more specific sub-concepts like the \texttt{cat's body}. 
Furthermore, in the third case of the third image, our SAE is capable of distinguishing between different individuals in a scene, whereas the vanilla SAE merely activates for the general category of ``person'' without individual-level differentiation.

Quantitatively, our proposed method consistently achieves a higher maximum IoU across all visualized cases, demonstrating that the structural regularizers successfully guide SAE neurons to align with meaningful, integrated semantic units rather than arbitrary low-level features.

\subsection{Interpreting Pipeline Design}
\label{sec:exp:pipline_design}

\begin{figure}[ht]
\centering
\includegraphics[width=\textwidth]{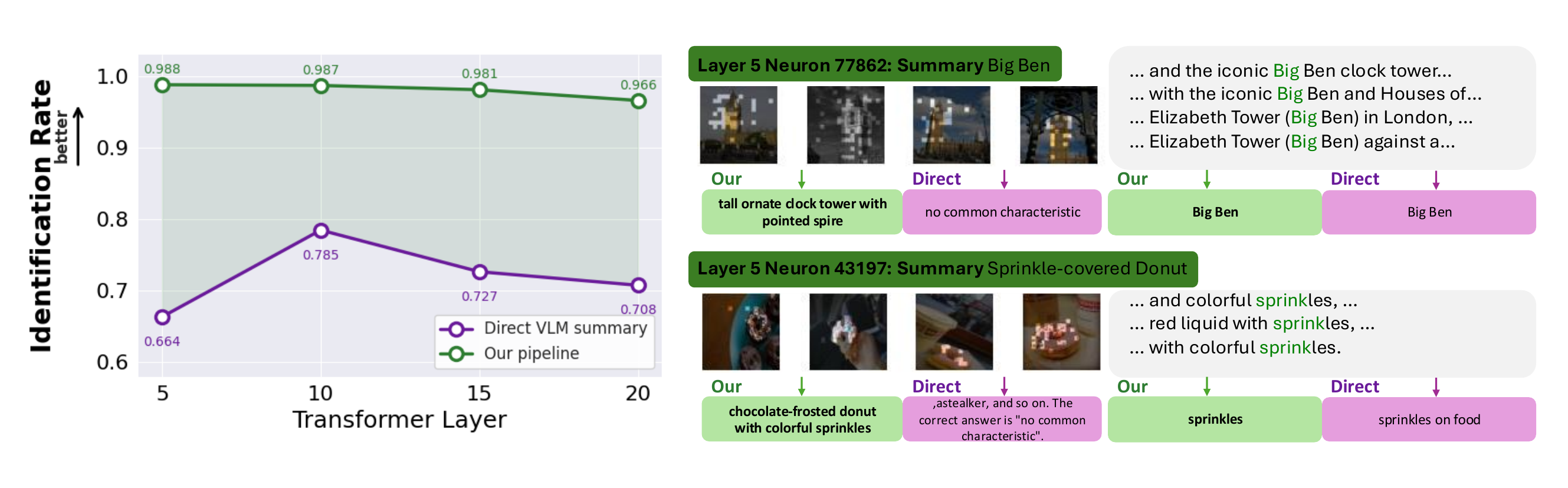}
\caption{
Comparison between hierarchical and direct interpretation pipelines.
\textbf{Left:} Identification rates across Transformer layers. 
Our hierarchical pipeline maintains near-perfect identification rates, while the direct summarization baseline is substantially less reliable. 
\textbf{Right:} SAE feature examples. 
Direct summarization often fails to extract common visual concepts from masked image references, whereas our pipeline first identifies local visual concepts and then summarizes them into coherent SAE feature descriptions.
}
\label{fig:pipeline_comparison}
\end{figure}

To validate the design of our automated interpretation pipeline, we compare our hierarchical summarization pipeline with a direct summarization baseline. 
Our pipeline first converts each activated visual reference into a concise textual description using a VLM (i.e., \texttt{Qwen3-VL-8B-Instruct}), and then uses an LLM (i.e., \texttt{Qwen3-30B-Instruct}) to summarize the common semantics across these image-derived descriptions. 
Similarly, for language references, the activated textual contexts are first interpreted into descriptions and then summarized into a language-side concept using the same LLM. 
In contrast, the direct baseline removes this intermediate description stage, i.e., it directly feeds the activated image references to the VLM to obtain a vision summary, and directly feeds the activated textual references to the VLM to obtain a language summary.
Here, to make the comparison as fair as possible, we use \texttt{Qwen3-VL-30B-Instruct} to align the number of parameters with the LLM we used in our pipeline. 

This comparison tests whether direct summarization is sufficiently reliable for SAE feature interpretation, or whether explicit concept identification before summary is necessary. 
As shown in \figureautorefname~\ref{fig:pipeline_comparison}, our hierarchical pipeline consistently achieves a much higher identification rate across all evaluated Transformer layers. 
This gap indicates that directly asking a VLM to infer a shared concept from multiple activated references is often unreliable, especially for visual references whose masks may contain partial objects, small discriminative regions, or spatially fragmented cues.
Furthermore, the SAE feature examples explain this failure mode. 
For Neuron 77862 of Layer 5, which corresponding to \texttt{Big Ben}, our pipeline identifies the visual concept as \textit{tall ornate clock tower with pointed spire}, while the direct baseline fails to extract a shared visual concept and outputs \textit{no common characteristic}. 
However, the language-side references clearly contain repeated mentions of \textit{Big Ben}, and both pipelines can summarize the language concept as \textit{Big Ben}. 
This suggests that direct summarization is not necessarily weak at recognizing explicit textual repetitions, but struggles when the concept must be inferred from masked visual evidence. 

These results justify the hierarchical design used in \sectionautorefname~\ref{sec:hierachical_pipeline}. 
Instead of relying on a VLM to directly perform both visual recognition and cross-reference abstraction, we decompose the interpretation task into two easier stages: localized concept identification and semantic summarization. 
This decomposition is especially important for the vision modality, where activated references are masked regions rather than complete natural images. 
By translating each visual activation into an explicit textual descriptor before summarization, our pipeline reduces visual ambiguity and yields more stable SAE feature interpretations.

\subsection{Clustering Design}
\label{sec:exp:clustering_design}

\begin{figure}[ht]
\centering
\includegraphics[width=\textwidth]{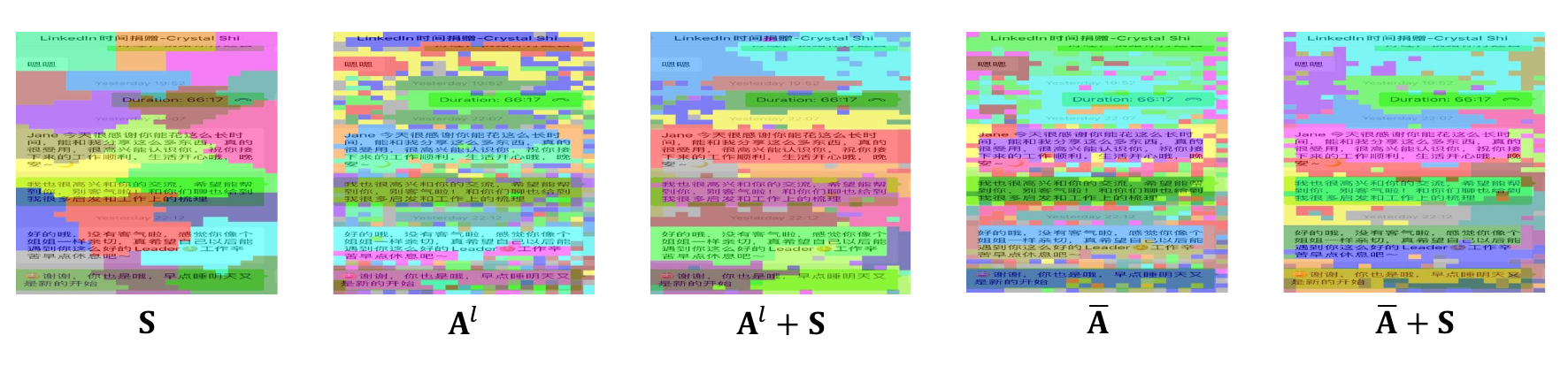}
\caption{
Visualization of clusters based on different adjacency criteria. 
\(\mathbf{S}\): spatial distance; \(\mathbf{A}^l\): attention on the current layer \(l\); \(\mathbf{\bar{A}}\): average attention over all layers.
\textbf{Takeaway:} \(\mathbf{\bar{A}} + \mathbf{S}\) yields best cohesive clusters with less intra-cluster noise. 
}
\label{fig:attn-all-or-next}
\end{figure}

This section discusses the choice of the adjacency matrix for clustering. 
The \figureautorefname~\ref{fig:attn-all-or-next} provides a visualization of clustering performance under various adjacency matrix configurations, illustrating their effectiveness in resolving semantic units. 
Utilizing the average attention over all layers ($\mathbf{\bar{A}}$) as the adjacency matrix yields more concentrated and cohesive clusters with significantly less intra-cluster noise compared to relying solely on the current layer’s attention ($\mathbf{A}^l$). 
Furthermore, the integration of spatial information ($\mathbf{S}$) provides an additional level of structural refinement, further enhancing the clarity and precision of the resulting clusters. 
Consequently, we adopt the combination of $\mathbf{\bar{A}} + \mathbf{S}$ as our final implementation to ensure optimal spatial-semantic alignment for SAE interpretability.

\begin{figure}[ht]
\centering
\includegraphics[width=.8\textwidth]{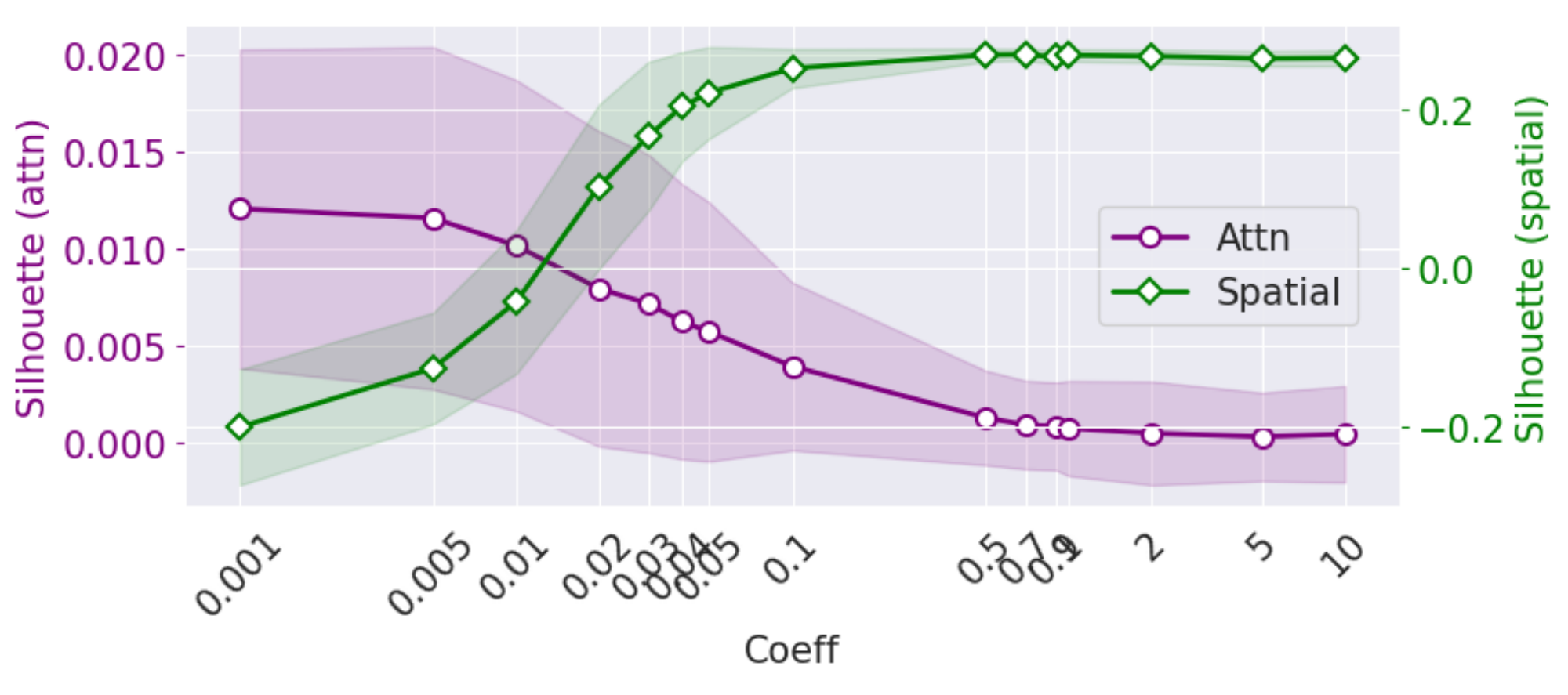}
\caption{
Impact of hyperparameter $\alpha$ on clustering quality. 
The plot illustrates the trade-off across a range of $\alpha$ values, with silhouette scores averaged over 50 samples. 
\textbf{Takeaway:} An optimal balance between these two metrics is achieved when $\alpha$ is positioned between 0.01 and 0.02. 
}
\label{fig:silhouette}
\end{figure}

Subsequently, to quantitatively evaluate the impact of the hyperparameter $\alpha$ on balancing semantic coherence and spatial continuity, we conduct a hyperparameter tuning experiment using the Silhouette Score~\citep{rousseeuw1987silhouettes} as the primary metric. 
The Silhouette Score is a measure of how similar an object is to its own cluster compared to other clusters. 
For a given sample $i$, the score is defined as $s(i) = \frac{b(i) - a(i)}{\max\{a(i), b(i)\}}$, where $a(i)$ represents the average distance between $i$ and all other points in the same cluster, and $b(i)$ denotes the average distance between $i$ and points in the nearest neighboring cluster. 
A score closer to 1 indicates that the sample is well-matched to its own cluster and poorly matched to neighboring clusters.
As illustrated in \figureautorefname~\ref{fig:silhouette}, we calculated Silhouette scores for two distinct adjacency matrices averaged over 50 samples: one based on attention weights ($\mathbf{\bar{A}}$), and another based on physical coordinates ($\mathbf{S}$). 
The plot reveals a clear trade-off: as the value of $\alpha$ increases, the silhouette score for $\mathbf{S}$ shows a significant upward trend before saturating, while the silhouette score for $\mathbf{\bar{A}}$ steadily declines.
This observation aligns with our underlying intuition. 
When $\alpha$ is small, the clustering process is primarily driven by attention weights; while this preserves strong semantic relationships, it often results in fragmented and spatially disjointed clusters due to the diffuse nature of visual features. 
Conversely, as $\alpha$ increases, the spatial distance ($\mathbf{S}$) becomes dominant, forcing adjacent patches to merge and thus improving spatial continuity. 
However, excessive spatial pressure can lead to the forced merging of semantically unrelated regions, compromising the overall semantic integrity of the clusters. 
Our results indicate that the two curves reach an optimal equilibrium within the range of $\alpha \in [0.01, 0.02]$. 
In this interval, the generated clusters effectively capture core semantic concepts while maintaining high spatial compactness. Consequently, we select $\alpha = 0.02$ as the default configuration for our system unless otherwise stated.

\subsection{Ablation Studies}


\begin{table}
    \centering
    \begin{tabular}{cccc}
    \toprule
        \textbf{ES} & \textbf{GS} & \textbf{$\text{mIOU}^c$} & \textbf{$\text{mIOU}^g$} \\
    \midrule
        \xmark & \xmark & 0.524 & 0.199 \\
        \cmark & \xmark & 0.568 & 0.080 \\
        \xmark & \cmark & 0.522 & 0.211 \\
        \cmark & \cmark & 0.555 & 0.153 \\
    \bottomrule
    \end{tabular}
    \caption{\textbf{Ablation studies} on exclusive sparsity (ES) and group sparsity (GS). Reported mIOU comparing active patch masks with cluster labels (\(\text{mIOU}^c\)) and within the same group (\(\text{mIOU}^g\)).}
    \label{table:ablation-esgs}
\end{table}

To evaluate the impact of our two structural regularizers, we conduct an ablation study focusing on Exclusive Sparsity (denoted as ES) and Group Sparsity (denoted as GS), which are introduced in \sectionautorefname~\ref{sec:esgs}. 
Before analyzing the results, we define Group Consistency ($\text{mIOU}^g$) as a metric to measure semantic coherence within a predefined group of patches. 
Specifically, for each group, we calculate the pairwise IoU of the SAE neuron activation patterns across all constituent patches. 
The final $\text{mIOU}^g$ is the average of these values across all groups, where a higher score indicates that patches within the same group consistently activate the same set of SAE neurons.
Specifically, $\text{mIOU}^c$ measures the spatial overlap between neuron activation masks and cluster labels in the patch space, whereas $\text{mIOU}^g$ evaluates the consistency of activation patterns across patches within the same group in the SAE neuron activation space.
As shown in \tableautorefname~\ref{table:ablation-esgs}, several key observations emerge:
\begin{itemize}[leftmargin=*,nosep]
    \item The application of ES (second row) significantly improves the cluster-level alignment ($\text{mIOU}^c$) (introduced in \sectionautorefname~\ref{sec:improved_vision_repr}) from 0.524 to 0.568. 
    This suggests that ES effectively encourages neurons to specialize in specific semantic clusters. 
    However, this comes at the cost of a sharp decline in group consistency ($\text{mIOU}^g$), which drops from 0.199 to 0.080, indicating that ES alone may lead to fragmented or inconsistent activations within a group.
    \item While GS alone (third row) provides a marginal improvement in group consistency ($\text{mIOU}^g=0.211$), its primary value is realized when combined with ES.
    \item By integrating both regularizers (fourth row), we achieve a balanced performance. 
    The inclusion of GS effectively counters the side effects of ES by restoring the group consistency to 0.153 while maintaining a high cluster-level alignment ($\text{mIOU}^c=0.555$).
\end{itemize}
In summary, the combination of ES and GS ensures that the learned features are not only discriminative at the cluster level but also semantically stable across related patches within a group.

\subsection{Visualization Website}
\begin{figure}[ht]
\centering
\includegraphics[width=\textwidth]{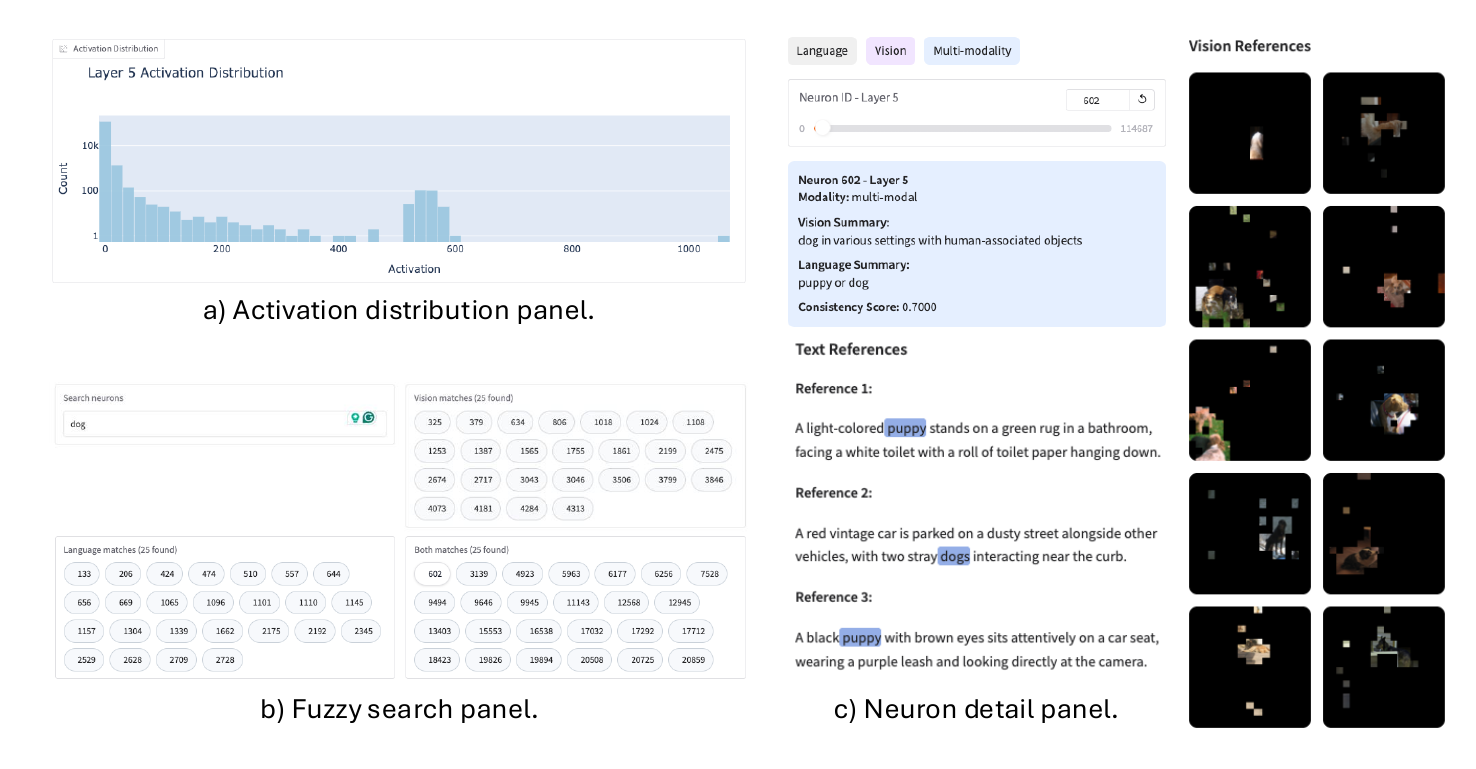}
\caption{
Example of messages shown on our website.
}
\label{fig:website}
\end{figure}

To further improve the usability of our proposed SAE and help the community to understand the details of the concepts, we developed a website\footnote{\href{https://huggingface.co/spaces/liaoweiduo/SAE-explorer}{https://huggingface.co/spaces/liaoweiduo/SAE-explorer}.} deployed in HuggingFace Spaces.  
As shown in \figureautorefname~\ref{fig:website}, this website provides modality statistics and an interactive SAE activation probability density distribution graph for each layer.
We also provide a fuzzy search panel to help users find specific neuron IDs related to the keywords in the vision and language summaries. 
Pressing the resulting pill buttons can directly relocate to the specific neurons. 
Next, the main panel, with a neuron ID scrolling bar, shows the details of the selected neuron. 
Specifically, we use a gallery view to show both the original image and the mask for vision references and highlight the active text tokens for text references. 

\section{Conclusion}

In this paper, we introduced the S$^2$AE, a novel framework designed to enhance the mechanistic interpretability of Vision-Language Models. 
Addressing the limitations of vanilla Top-K SAEs, which often suffer from polysemantic noise and a lack of spatial coherence in the visual domain, we proposed a visual region consistency regularizer that leverages both Transformer attention similarity and spatial proximity to cluster patches into semantically meaningful units. 
By incorporating structured sparsity regularization, specifically through the combination of exclusive sparsity for inter-region disentanglement and group sparsity for intra-region consistency, S$^2$AE successfully encourages SAE features to specialize in distinct, coherent visual concepts.
We develop on the \texttt{Qwen2.5-VL-7B-Instruct} model and demonstrate that S$^2$AE achieves superior semantic alignment (measured by \(\text{mIoU}^c\) and visualized ) and representational efficiency (lower \(\ell_0\) norm) while maintaining near-perfect reconstruction fidelity with an Explained Variance above 0.99. 
We also validate the design of our automated interpretation pipeline, showing that hierarchical concept identification followed by semantic synthesis is substantially more reliable than direct VLM summarization for masked SAE references.
More importantly, our results show that visual structure is not merely a local regularizer for image patches. 
Because multi-modal SAEs use a shared feature dictionary for vision and language tokens, cleaner visual concept decomposition can propagate to multi-modal features, improving their language-side monosemanticity and cross-modal consistency. 
This finding broadens the role of visual structural priors from improving visual interpretability to organizing shared multi-modal concept representations.


\clearpage
\bibliography{main}

\appendix

\section{Implementation Details}
\label{sec:impl_details}

\subsection{Implementation Details}

We train SAE with hidden residuals extracted from \texttt{Qwen2.5-VL-7B-Instruct}'s layer \{5,10,15,20\}. 
The training data is two-fold: images are provided by~\citet{zhang2025large} and the corresponding textual descriptions are produced by \texttt{Qwen3-VL-8B-Instruct}, described in \sectionautorefname~\ref{sec:reference-source-construction}. 
The hyperparameters are summarized in \tableautorefname~\ref{table:hyperparameters}.
We use 4 NVIDIA A800 GPUs to extract hidden residuals and train our SAE. 
When collecting reference samples, we store top-10 samples for each modality. 
For vision reference identification, we use \texttt{Qwen3-VL-8B-Instruct} to identify the concepts indicated by masks. 
For language reference identification, we use \texttt{Qwen3-30B-Instruct} to identify the concepts indicated by the highlighted tokens. 
For concept summarization and modality consistency scoring, we use \texttt{Qwen3-30B-Instruct}. 

\begin{table}[htbp]
  \centering
  \caption{Hyperparameters.}
  \label{table:hyperparameters}
  \begin{tabular}{c|c}
    \toprule
    Hyper-parameters & Value \\
    \midrule
    \(N\) & 114688 \\
    \(D\) & 3584 \\
    \(k\) & 256 \\
    \midrule
    \(\lambda_{\rm{aux}}\) & 0.1 \\
    \(G\) & 20 \\
    \(\alpha\) & 0.02 \\
    \(\lambda_{\rm{es}}\) & 0.1 \\
    \(\lambda_{\rm{gs}}\) & 0.01 \\
    \bottomrule
  \end{tabular}
\end{table}

\subsection{Prompt Details}

The prompts for visual and language concept identification are shown in Prompt~\ref{prompt:image_identification} and Prompt~\ref{prompt:text_identification}, respectively.
The prompt for concept summarization is shown in Prompt~\ref{prompt:summarization}. 
The prompt for obtaining modality consistency is shown in Prompt~\ref{prompt:consistency}. 

\begin{promptbox}[prompt:image_identification]{Visual Concept Identification}
    \small
    \medskip
    
    \textbf{[SYSTEM]} \\
    You are a meticulous AI researcher conducting an important investigation into the behavior of an image marker. \\
    Your task is to analyze which visual feature or concept is indicated by a mask (highlighted) region from a specific image and provide a brief explanation that encapsulates its behavior. \\
    
    \medskip
    \textbf{[REQUIREMENTS]} \\
    \begin{enumerate}[leftmargin=*,nosep]
        \item Focus only on the highlighted region in the masked image. If no region is highlighted (i.e., the given mask is empty),  output: ``[EXPLANATION]: Unable to produce descriptions.'' 
        \item If the highlighted region is minimal (e.g., a few bright spots), you should first consider whether it is one part of a larger object or background, like sky or wall. If it is not or indicates noisy spots, output: ``[EXPLANATION]: Unable to produce descriptions.'' 
        \item The original image is provided for reference, but your analysis and explanation should be based solely on the highlighted region in the masked image. 
        \item Identify common visual patterns, objects, or concepts in the activated regions. For example, note if highlighted areas show structures, such as mesh patterns or concrete objects. 
        \item If the activated regions depict abstract concepts like human actions or emotions, describe these actions or emotions directly, such as a happy face, or a visual distortion characteristic of a fisheye lens. 
    \end{enumerate}

    \medskip
    \textbf{[GUIDELINES]} \\
    You will first receive an image and then receive the corresponding masked image where specific regions have been highlighted. \\
    Non-highlighted areas will be masked out or dimmed. \\
    Your analysis should consider only the highlighted regions on the masked image and complete the following tasks: \\
    \begin{enumerate}[leftmargin=*,nosep]
        \item \textbf{Describe Only the Highlighted Regions:} Generate captions solely based on the highlighted regions, specifically in the masked image. If no meaningful pattern is visible, or if only a few scattered noisy spots are highlighted, output: ``[EXPLANATION]: Unable to produce descriptions.'' 
        \item \textbf{Concise Description Only:} Provide a short, direct description of the common features within the highlighted regions from the provided images. Avoid any interpretive language-simply state what you see, such as ``mesh-like structures'' or ``actions related to joy or happiness.'' 
        \item \textbf{Output Format:} Ensure beginning the response with ``[EXPLANATION]:'' followed by your explanation, if applicable. If unable to determine common visual features, output: ``[EXPLANATION]: Unable to produce descriptions.'' 
    \end{enumerate}

    \medskip
    \textbf{[TASK CONTENT]} \\
    The original image is: \{\{ORIGINAL IMAGE\}\}. \\
    The masked image is: \{\{MASKED IMAGE\}\}.
\end{promptbox}

\begin{promptbox}[prompt:text_identification]{Language Concept Identification}
    \small
    \medskip
    
    \textbf{[SYSTEM]} \\
    You are a meticulous AI researcher investigating which textual feature or concept is indicated by highlighted spans in a sentence. \\
    
    \medskip
    \textbf{[REQUIREMENTS]} \\
    \begin{enumerate}[leftmargin=*,nosep]
    \item Focus only on text enclosed by \texttt{<<} and \texttt{>>}. If no span is highlighted or spans are empty/noisy, output: ``[EXPLANATION]: Unable to produce descriptions.''.
    \item Provide a concise description of the concept or feature implied by the highlighted span(s). Keep it short and direct.
    \item Begin the response with ``[EXPLANATION]:''. If nothing meaningful can be inferred, output exactly ``[EXPLANATION]: Unable to produce descriptions.''.
    \end{enumerate}

    \medskip
    \textbf{[EXAMPLES]} \\ 
    Example 1: \\
    Input:  \\
    A lively beach gathering features people relaxing under colorful umbrellas and tents, with some cooking over a grill and others social\texttt{<<}izing\texttt{>>} in beach chairs. \\
    Output: \\
    \text{[EXPLANATION]:} social interaction \\
    
    Example 2: \\
    Input:  \\
    A group of people, including an elderly man in the \texttt{<<}foreground\texttt{>>} and a smiling woman beside him, are enjoying a festive meal together at a crowded restaurant table adorned with food, drinks, and holiday decorations. \\
    Output: \\
    \text{[EXPLANATION]:} foreground \\
    
    \medskip
    \textbf{[TASK CONTENT]} \\
    Below is the text with highlighted spans delimited by \texttt{<<} and \texttt{>>}. \\
    \{\{MASKED TEXT\}\}.
\end{promptbox}

\begin{promptbox}[prompt:summarization]{Concept Summarization}
    \small
    \medskip
    \textbf{[SYSTEM]} \\
    You are an AI assistant assessing if descriptions consistently depict the same patterns, objects, or concepts. \\

    \medskip
    \textbf{[GUIDELINES]} \\
    \textbf{Input:} Descriptions each starting with ``[EXPLANATION]:''. \\
    \textbf{Output rules:} \\
    \begin{enumerate}[leftmargin=*,nosep]
        \item \textbf{If all align:} Output ONLY a short concept phrase (noun phrase), not a sentence. Examples: ``mesh-like structure'', ``happy face'', ``fisheye lens distortion''.
        \item Ignore \(\leq\)2 deviations or ``Unable to produce descriptions''; summarize the majority. 
        \item If no clear commonality or \(>\)half are ``Unable to produce descriptions'': Output ``no common characteristic''. 
    \end{enumerate}
    
    Do NOT add any prefix like ``[EXPLANATION]:''. Output ONLY the concept phrase or ``no common characteristic''. \\

    \medskip
    \textbf{[EXAMPLES]} \\
    Example 1:   \\
    Input descriptions:  \\
    \text{[EXPLANATION]:} A small portion of a giraffe's leg and the adjacent green grass are visible in the highlighted region. \\
    \text{[EXPLANATION]:} A close-up of a human eye with visible eyelid and surrounding skin texture. \\
    \text{[EXPLANATION]:} Close-up of a human eye and adjacent skin with visible wrinkles and fine texture. \\
    \text{[EXPLANATION]:} A person's face with wide, surprised eyes and an open mouth, indicating a startled or shocked expression. \\
    \text{[EXPLANATION]:} A person's face with wide, surprised eyes and an open mouth, conveying an expression of shock or astonishment. \\
    \text{[EXPLANATION]:} Unable to produce descriptions. \\
    \text{[EXPLANATION]:} Unable to produce descriptions. \\
    \text{[EXPLANATION]:} Unable to produce descriptions. \\
    \text{[EXPLANATION]:} Unable to produce descriptions. \\
    \text{[EXPLANATION]:} A dense cluster of dry, golden-brown straw stalks with varied orientations and textures, forming a fibrous, interwoven ground cover. \\
    Output commonality: \\
    human eye with expressive features \\
    
    Example 2: \\
    Input descriptions:\\
    \text{[EXPLANATION]:} Unable to produce descriptions.\\
    \text{[EXPLANATION]:} Unable to produce descriptions.\\
    \text{[EXPLANATION]:} Unable to produce descriptions.\\
    \text{[EXPLANATION]:} Unable to produce descriptions.\\
    \text{[EXPLANATION]:} Unable to produce descriptions.\\
    \text{[EXPLANATION]:} The highlighted regions show the top ornate edge of a clock face and a partial view of a banner with the word \"GREY\".\\
    \text{[EXPLANATION]:} Textual branding on umbrella panels, including visible logos and typography, amid falling snow.\\
    \text{[EXPLANATION]:} Unable to produce descriptions.\\
    \text{[EXPLANATION]:} A solid, vertically oriented rectangular region with a uniform golden-yellow hue, suggesting a flat, possibly textured surface like a book cover or framed artwork.\\
    \text{[EXPLANATION]:} A protest sign with handwritten text reading \"It's a Pink Slip\" and a symbol of equality (two parallel lines with an equals sign).\\
    Output commonality:\\
    no common characteristic\\
    
    Example 3: \\
    Input descriptions: \\
    \text{[EXPLANATION]:}  socializing in\\
    \text{[EXPLANATION]:}  socializing and\\
    \text{[EXPLANATION]:}  socializing and\\
    \text{[EXPLANATION]:}  socializing around\\
    \text{[EXPLANATION]:}  socializing at\\
    \text{[EXPLANATION]:}  socializing around\\
    \text{[EXPLANATION]:}  socializing around\\
    \text{[EXPLANATION]:}  socializing at\\
    \text{[EXPLANATION]:}  socialize.\\
    \text{[EXPLANATION]:}  socializing in\\
    Output commonality:\\
    social interaction\\
    
    \medskip 
    \textbf{[TASK CONTENT]} \\
    Here are the descriptions: \\
    \{\{A SEQUENCE OF DESCRIPTIONS\}\}
\end{promptbox}

\begin{promptbox}[prompt:consistency]{Modality Consistency}
    \small
    \medskip
    \textbf{[SYSTEM]} \\
    You are an AI assistant scoring how well two descriptions refer to the same underlying pattern/object/concept. \\

    \medskip
    \textbf{[REQUIREMENTS]} \\
    You should:
    \begin{itemize}[leftmargin=*,nosep]
        \item Output a single number in \([0,1]\) with one decimal place (e.g., 0.0, 0.5, 1.0).
        \item If any description is ``no common characteristic'' or indicates inability to describe, output 0.0.
        \item Exact match or one is a clear specific instance of the other \(\rightarrow\) 1.0. (dog species vs dog \(\rightarrow\) 1.0)
        \item Strong partial/part-whole relation \(\rightarrow\) around 0.7. (dog vs dog tongue \(\rightarrow\) 0.7)
        \item Same broad category but different subtypes \(\rightarrow\) around 0.5. (chihuahua vs malamute \(\rightarrow\) 0.5)
        \item Different or conflicting concepts \(\rightarrow\) 0.0. (animal vs plants \(\rightarrow\) 0.0)
        \item Be conservative; do not output text, only the score.
    \end{itemize}

    \medskip
    \textbf{[EXAMPLES]} \\
    Example 1: \\
    Input:\\
    Description 1: dog species \\
    Description 2: dog \\
    Output: \\
    1.0 \\
    
    Example 2: \\
    Input:\\
    Description 1: dog \\
    Description 2: dog tongue \\
    Output: \\
    0.7 \\
    
    Example 3: \\
    Input: \\
    Description 1: chihuahua \\
    Description 2: malamute \\
    Output: \\
    0.5 \\
    
    Example 4: \\
    Input: \\
    Description 1: animal \\
    Description 2: plants \\
    Output: \\
    0.0 \\
    
    \medskip
    \textbf{[TASK CONTENT]} \\
    Here are the descriptions: \\
    Description 1: \{\{VISION DESCRIPTIONS\}\} \\
    Description 2: \{\{LANGUAGE DESCRIPTIONS\}\} \\

\end{promptbox}

\appendix
\end{document}